\documentclass[a4paper,fleqn]{cas-dc}

\usepackage[numbers]{natbib}
\usepackage{amsmath}
\usepackage{algorithm}
\usepackage{algpseudocode}
\usepackage{balance}

\def\tsc#1{\csdef{#1}{\textsc{\lowercase{#1}}\xspace}}
\tsc{WGM}
\tsc{QE}

\begin{document}
\let\WriteBookmarks\relax
\def\floatpagepagefraction{1}
\def\textpagefraction{.001}

\shorttitle{An Ergonomic Role Allocation Framework for Dynamic Human-Robot Collaborative Tasks}

\shortauthors{E. Merlo, E. Lamon et al.}  
\title [mode = title]{An Ergonomic Role Allocation Framework for Dynamic Human-Robot Collaborative Tasks} 

\author[1,2]{Elena Merlo}[]
\cormark[1]
\author[1]{Edoardo Lamon}[]
\cormark[1]
\ead{edoardo.lamon@iit.it}
\author[1,3]{Fabio Fusaro}[]
\author[1]{Marta Lorenzini}[]
\author[2]{Alessandro Carfì}[] 
\author[2]{Fulvio Mastrogiovanni}[] 
\author[1]{Arash Ajoudani}[] 

\affiliation[1]{organization={Human-Robot Interfaces and Interaction, Istituto Italiano di Tecnologia},
            city={Genoa},
            country={Italy}
}

\affiliation[2]{organization={Dept. of Informatics, Bioengineering, Robotics, and Systems Engineering, University of Genoa},
            city={Genoa},
            country={Italy}
}
            
\affiliation[3]{organization={Dept. of Electronics, Information and Bioengineering, Politecnico di Milano},
            city={Milan},
            country={Italy}
}

\cortext[1]{Contributed equally. Corresponding author:}

\begin{abstract}
By incorporating ergonomics principles into the task allocation processes, human-robot collaboration (HRC) frameworks can favour the prevention of work-related musculoskeletal disorders (WMSDs). In this context, existing offline methodologies do not account for the variability of human actions and states; therefore, planning and dynamically assigning roles in human-robot teams remains an unaddressed challenge.
This study aims to create an ergonomic role allocation framework that optimises the HRC, taking into account task features and human state measurements. The presented framework consists of two main modules: the first provides the HRC task model, exploiting AND/OR Graphs (AOG)s, which we adapted to solve the allocation problem; the second module describes the ergonomic risk assessment during task execution through a risk indicator and updates the AOG-related variables to influence future task allocation. 
The proposed framework can be combined with any time-varying ergonomic risk indicator that evaluates human cognitive and physical burden. In this work, we tested our framework in an assembly scenario, introducing a risk index named Kinematic Wear.
The overall framework has been tested with a multi-subject experiment. The task allocation results and subjective evaluations, measured with questionnaires, show that high-risk actions are correctly recognised and not assigned to humans, reducing fatigue and frustration in collaborative tasks.
\end{abstract}


\maketitle

\section{Introduction}
\label{sec:intro}

The research field studying collaborative processes, in which humans and robots work together to achieve shared goals, is known as Human-Robot Collaboration (HRC).
HRC can improve industrial manufacturing and logistic processes by exploiting humans' and robots' skills at the same time \cite{ajoudani2018progress,fitts1951human}.
Due to their easy and fast reconfigurability, robotic platforms can meet the requirements of the new products' demands, appearing practical in small and medium-sized enterprises (SMEs), characterised by flexible and varying production lines. 
Furthermore, HRC frameworks may be empowered to embed safety \cite{haddadin2009requirements,desantis2008atlas}, and ergonomics \cite{maurice2017human, kim2019adaptable} principles. This potentially responds to the urgency of reducing work-related musculoskeletal disorders (WMSDs), which are the primary cause of absenteeism and lost productivity in industries \cite{govaerts2021prevalence}.

To provide robots with information about workers' ergonomic risk, the careful monitoring of the biomechanical risk associated with their activities is fundamental. 
For this purpose, numerous approaches can be found in literature. Most of them analyses a particular aspect or a specific activity: carrying and lifting \cite{waters1994applications}, pushing and pulling \cite{snook1991design}, low loads at high frequency \cite{occhipinti1998ocra}, posture and movements \cite{hignett2000rapid,karhu1981observing,mcatamney1993rula}.
Within HRC frameworks, to assess the ergonomic risk during hybrid task execution, mainly posture-based indexes have been exploited \cite{busch2017postural, faber2017cognition, van2018improving}. The latter evaluate the body configurations that the human is adopting while performing the task and assign a bio-mechanical risk to the actions based on the range of motion of the joints involved and the type of movements (i.e., bending, twisting, etc.). However, the choice of the most suitable ergonomic index to drive the robot behaviour should be driven by the task features. For instance, if the human agent is required to manipulate heavy objects, a posture-based index cannot describe the associated risk. In some cases, the information about human dynamics (i.e., moments and forces developed within the human body) is the key requirement for the robot adaptation \cite{peternel2017human, kim2017anticipatory, parastegari2017modeling, marin2018optimizing, lorenzini2019new}.
On the other hand, the digitalisation of the workplace may lead to work intensification and constant time pressure, calling for the design of indexes that can assess the workers' mental stress and extra cognitive load \cite{rajavenkatanarayanan2020towards, lagomarsino2021online}. 

\begin{figure*} [!t]
    \centering
	\includegraphics[trim=0.0cm 0.0cm 0.0cm 0.0cm,clip,width=1\linewidth]{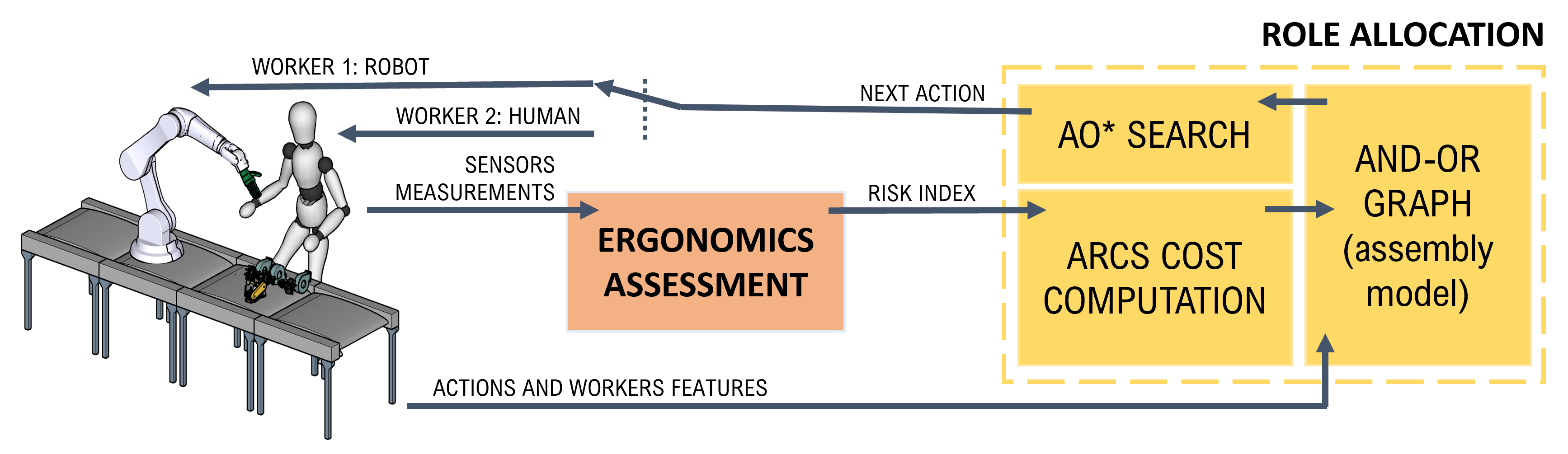}
	\vspace{-4mm}
    \caption{The framework scheme emphasises the distinction between its two main modules. The "Ergonomics Assessment" module, during the human action performance, is responsible for the human state evaluation, updating the ergonomic risk that will change the cost of each AOG arc. The "Role Allocation" provides at each iteration the next task-allocated worker pairing, which is the result of the AO* search through the AOG structure, modelled accounting for task features and workers’ conditions.}
	\label{fig:framework}
	\vspace{-0.3cm}
\end{figure*}

Given the advantages of collaborative solutions for workcell settings, 
researchers have tried to close the gap between the industrial requirements and the HRC paradigm by focusing on human-robot interfaces, robot control modalities, robot intelligence, and robot perception of the human psycho-physical state to organise an appropriate collaboration plan.
One of the main issues in defining a teamwork strategy is how to distribute the roles among teammates. Literature refers to it as the \textit{role allocation} problem. Role allocation is mostly studied within the context of multi-robot systems \cite{farinelli2004multirobot}, where team coordination is necessary to accomplish the final common objective. The typical approach consists of designing functions, often called \textit{utilities}, able to describe the quality of agent-task pairing and find the optimal solution through an optimisation algorithm. 
Calculating an agent's utility is not trivial because it depends on the agent's aptitude for the role's responsibilities, the role's priority, and the team's current and future needs.
Generalising a multi-robot scenario to a mixed team of agents (robots and humans) requires defining a utility function for human workers that is consistent with, i.e., comparable to, the utility functions indicating whether robots are suitable for specific roles. Examples in literature suggest that such a function can consider, among many, the time spent to complete the activity \cite{pearce2018optimizing, fusaro2021human}, the travelled distance if the human is required to move from one workstation to another \cite{banziger2018optimizing}, the perceived effort for the task completion \cite{pearce2018optimizing}.

One of the methods employed for modelling the collaborative task accounting for task features is the AND/OR Graph (AOG) structure \cite{demello1990and}.
AOG has been introduced for describing well-structured tasks and its decomposition into a sequence of actions, and it is an efficient representation of a state transition graph, with few nodes, that allows an easy search for feasible plans.
The authors of \cite{johannsmeier2016hierarchical, darvish2018flexible} recently
proposed two different offline optimisation methods based on AOG to solve simultaneously the planning and the role allocation problems.
In addition, three quantifiable indexes were suggested in \cite{lamon2019capability} to more accurately describe the kinematic and dynamic properties of the agents, resulting in a capability-aware allocation.
The three indexes were combined by employing a weighted sum, obtaining a proper heuristic for the optimisation algorithm.
Other methods addressing an ergonomic role allocation were proposed by \cite{el2019task, el2022hierarchical}.

However, one of the main limitations of AOG-based approaches is that they provide the role distribution beforehand, i.e., actions are allocated to agents before starting the collaboration. As a result, the allocation remains fixed for the entire teamwork, making it difficult to account for changing human conditions.
In light of these considerations, this manuscript aims to:
\begin{enumerate}[i.]
    \item design an HRC AOG-based strategy, accounting for a dynamic role allocation, i.e., where the roles are not fixed within the team but change on-the-fly according to a specific optimisation criteria, and
    \item capitalise on human ergonomic risk model to evaluate the suitability of an agent to the action, in order to prevent the assignment of potentially risky actions to human workers.
\end{enumerate}
In particular, a human-monitoring system is employed to evaluate online the ergonomic state of the worker, profiting from measurements of human physical or cognitive workload.
The human monitoring system is mainly used to perform the ergonomic assessment of the human worker, while it does not contribute actively to action recognition, which is out of the scope of the paper. Thus, we assume that the worker will execute the assigned action and a mechanism, to understand when the action has been completed, is available.
To integrate the ergonomic constraints into the AOG structure, the proposed framework i) updates the AOG arc costs on-the-fly according to the human  ergonomic risk model and ii) uses an AO* search to find iteratively the optimal action-agent pairing.
In \autoref{fig:framework} a scheme of the designed framework is shown.

We tested the functioning of our online human-robot role allocation method in a simple assembly of lightweight objects. Since existing standard methods provide an overall rough evaluation of the task and do not account for the history of body postures, as an ergonomic indicator, we propose a custom index, named Kinematic Wear (KWear), that embodies the subject \textit{ergonomic history}, memorising the usage of each joint during the whole execution of an assembly task. However, the proposed framework can also be combined with other metrics that describe human risk.

A preliminary version of this work is presented in \cite{merlo2022dynamic} which only sketched the ergonomic role allocation framework. In this work, the details of the aforementioned framework are outlined, adding novel extensive performance testing and validation in a realistic scenario through multi-subject evaluation. In particular, with respect to \cite{merlo2022dynamic}, the novel contributions are the following:
\begin{enumerate}[i.]
    \item an improved dynamic role allocation strategy based on AOGs with continuous hyper-arc cost functions;
    \item a computational complexity investigation in case of scarcely and sequentially connected assemblies;
    \item an analysis of the suitability of KWear as metric for the role allocation framework in predicting future ergonomic risk, including a comparison with a static ergonomic metric (RULA); 
    \item a subjective evaluation of the usability of our framework in industrial contexts.
\end{enumerate}

The paper proceeds by describing in detail each framework block. In \autoref{sec:AOG}, we explain how the AOG structure models a collaborative task embedding workers' features, and in \autoref{sec:AO*}, we describe the search algorithm AO* that we implemented for inspecting the graph at each iteration. In \autoref{sec:cost_comp}, we illustrate the rules used for mapping the risk index into hyper-arcs cost, and in \autoref{sec:erg_assess} we focus on how ergonomics can be integrated into the proposed framework and on the design of the KWear index. In \autoref{sec:comp_compl}, the computational complexity analysis of our role allocation algorithm is presented. In \autoref{sec:expe}, we outline experimental analysis conducted for testing KWear properties first, and then for assessing the performance of the overall framework with a proof-of-concept assembly involving 12 subjects. Finally, future research directions are presented in \autoref{sec:concl}.

\section{AND/OR Graphs for Role and Task Planning}
\label{sec:AOG}
One of the approaches for representing well-structured industrial tasks, such as assemblies, exploits AOGs. AOGs are data structures able to model all the possible combination sequences of an assembly task (or any sequential task in general) in a compact representation. 

\subsection{Assembly Modelling}
AOGs describe compactly all possible assembly sequences for a certain product, which is composed of different atomic pieces. The advantage of using such graphs with respect to other graph-based strategies, such as the Assembly State Graphs, lies in having fewer nodes and a simpler search for the optimal assembly plan guided by the weights associated with each arc in the graph~\cite{demello1990and}. 

\begin{figure}
    \centering
    \includegraphics[scale=0.5]{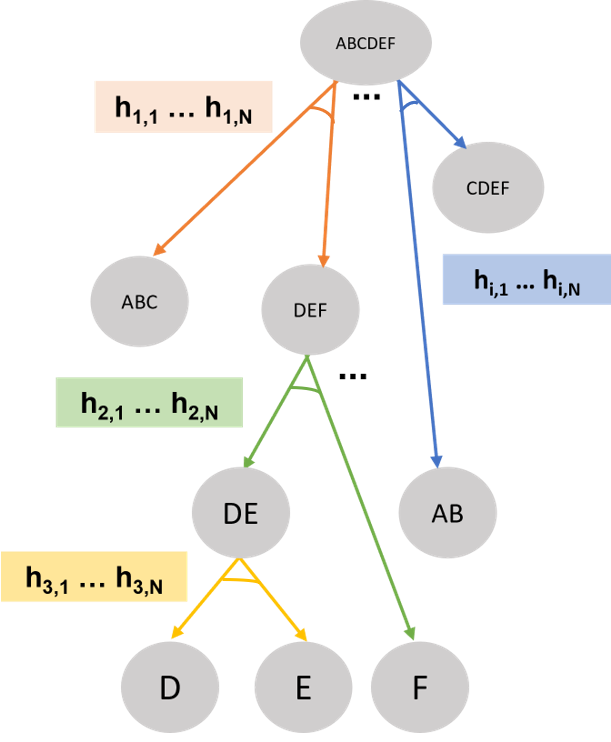}
    \caption{General AOG of an assembly made up of six pieces: A, B, C, D, E and F. Each hyper-arc has a different color since it represents a different assembly action. Each hyper-arc is duplicated to represent the feasibility of the action by each agent. Each arc has a different cost according to the suitability of each agent for performing the corresponding action.}
    \label{fig:aog_new}
    \vspace{-2mm}
\end{figure}

Given an assembly task $P$ made of $M$ pieces ($P = \{ p_1, p_2, \dots, p_M \}$), a configuration $\Theta$ of $P$ is a set of sub-assemblies of $P$ such that:
\begin{itemize}
    \item each sub-assembly in $\Theta$ is formed by physically feasible and stable connections;
    \item each $p_i$ belongs to one sub-assembly of $\Theta$;
    \item a $p_i$ cannot belong to two or more sub-assembly of $\Theta$.
\end{itemize}
The assembly plan can be seen as a sequence of $\Theta$s, starting from $\Theta_i = \{\{ p_1 \},\{ p_2 \},\dots,\{ p_M\}\}$ and ending with $\Theta_f = \{\{ p_1, p_2, \dots, p_M\}\}$, i.e., starting from the configuration in which all the atomic pieces are separated (each of them is a sub-assembly), and ending with a unique final subset, corresponding to a configuration with all the pieces assembled.
By considering feasible sub-assemblies as nodes and assembly operations that describe the transition between two configurations as edges, it is possible to find a path from $\Theta_i$ to $\Theta_f$ using path search algorithms.
The edges, called \textit{hyper-arcs} are pairs in which:
\begin{itemize}
    \item the first element is a single node (father);
    \item the second element is a set of nodes (children); children represent all the possible sub-assemblies that can be obtained by disassembling the father node. For assembly tasks, since most of the assembly operations consist of joining two sub-assemblies, hyper-arcs are modelled as two-to-one connectors.
\end{itemize}
Thus, an AOG is described by a set of nodes $N=\{n_1, n_2, \dots,$ $ n_{|N|}\}$,
and a set of hyper-arcs $H=\{h_1, h_2, \dots, h_{|H|}\}$.
Each node $n \in N$ represents a sub-assembly of $P$, while hyper-arcs define the assembly operations and can be characterised by different costs~\cite{chang1971admissible}. 
Children connected by the same $h$ are in a logical AND, while different hyper-arcs with the same parent node are in a logical OR.
In general, it is possible to obtain a specific assembly configuration $\bar{\Theta}$ performing assembly operations between different sub-assemblies. Such sub-assemblies are in AND relation, while the different assembly operations are represented by hyper-arcs in OR relation. 
For example, as described in \autoref{fig:aog_new}, to complete the assembly $ABCDEF$, it is possible to join $ABC$ with $DEF$ (assembly operation described by hyper-arc $h_1$) or join $AB$ with $CDEF$ represented by $h_i$. Thus, $ABC$ is linked in AND with $DEF$ and $AB$ with $CDEF$, while $h_1$ and $h_i$ are in a logical OR.
The only node without a father is named root. The nodes without children are identified as leaf nodes, and they are as many as the assembly pieces.

\subsection{Role Allocation}

The goal of this study is to provide a formulation of AOGs that could provide not only the sequence of assembly operations but also the agent in charge of such assembly. To do so, we extend the standard AOG definition with two supplementary sets: the set of workers involved in the task execution $W=\{w_1, w_2, \dots, w_{|W|}\}$ and the set of actions that have to be performed, $A=\{a_1, a_2, \dots, a_{|A|}\}$. For generality, actions represent proper assemblies (e.g., interlocking two parts together) and \textit{relaxed} assemblies (e.g., moving an object on top of a table could be considered an assembly between such an object and the table).
The assembly sequence allocated to each worker is obtained by duplicating the same hyper-arc (that describe an assembly operation) for $|W|$ times and assigning a cost to each $h_i \in H$, i.e. $c_{h_i, w_j}$ with $i\in[1,|H|]$ and $j\in[1,|W|]$, that represents the suitability of $w_j$ in performing such assembly operation. If a specific action cannot be executed by one of the involved agent, or results as unsafe or time-consuming, to prevent the agent to be assigned to the action, we prune the corresponding hyper-arc. 
Finally, by exploiting an optimality-based search algorithm, the path with the minimum total cost can be found (\autoref{sec:AO*}).
Unlike in \cite{johannsmeier2016hierarchical}, where costs were fixed for the whole duration of the assembly task, 
in our framework costs are updated according to the human state monitored during the cooperation, ensuring an optimal role assignment at each assembly step. The description of the costs update mechanism and the monitored variables can be found in \autoref{sec:cost_comp} and \autoref{sec:erg_assess}, respectively.

\section{AO* for the Optimal Search}
\label{sec:AO*}
A custom AO* search is implemented to retrieve the desired assembly sequence and the optimal action allocation. Such an algorithm is inspired by the A*~\cite{hart1968formal}, which is adapted to operate on an AOG.
Unlike standard AO*, whose objective is to minimise the path cost from the root to a single leaf or a pair of leaf nodes \cite{martelli1973additive},  the goal of our algorithm is to inspect leaf nodes, as they represent the initial configuration $\Theta_i$ of the assembly. In this way, the search of our AO* does not finish until all the leaf nodes, i.e., all the assembly pieces in $\Theta_i$, are visited and included in the path.

The algorithm exploits an auxiliary tree data structure $(Z, E)$ that is updated during the search and contains the explored assemblies at minimum cost. $Z$ is the set of nodes of the tree, where each node consists of a \textit{SearchState}. 
Each \textit{SearchState} has the following attributes:
(i) an assembly configuration $\Theta$, that represents the state of the assembly at the current step of the search algorithm;
(ii) a pointer to its father assembly configuration, that represents the state of the assembly at the previous step of the search algorithm;
(iii) a score $c$, that represents the cost of reaching $\Theta$.
$E$ is the set of edges that connects a \textit{SearchState} containing an assembly configuration at level $l$ with its father at level $l$-$1$. A cost is associated with each edge, that corresponds to the cost of the transition between $\Theta$ and its father. Such a cost is inferred by the AOG hyper-arc connecting the sub-assemblies in $\Theta$ with the resultant sub-assembly in its father. 

\alglanguage{pseudocode}
\begin{algorithm} [!t]
\small
\caption{AO* search}\label{algo:ao*}
\begin{algorithmic}[1]
\Require{$\Theta_f$, $\Theta_i$ }{}
\State Path \textit{opt\_path}
\State {SearchState \textit{start}, \textit{goal}}
\State {\textit{start}.setState($\Theta_f$)}
\State {\textit{goal}.setState($\Theta_i$)}
\State {\textproc{add}(\textit{start} \textproc{in} OPEN)}
\While{$\neg\,$(OPEN.empty())}
    \State {\textit{curr} $\gets$ \textproc{min}($\Theta.c$)}
    \If{\textit{curr} $==$ \textit{goal}}
        \State \textbf{break}
    \EndIf
    \For{$\theta_j$ \textproc{in} $\Theta_{curr}$ }
        \For{edge $i$ \textproc{connected} to $j$}
            \State {SearchState \textit{succ}}
            \State {\textit{succ} $\gets$ \textit{curr}}
            \State {$\Theta_{curr}$ $\gets$ $\theta$ \textproc{pointed} by $i$}
            \State \textit{succ}.father $\gets$ \textit{curr}
            \State \textit{succ}$.c$ $\gets$ \textit{curr}$.c$ + \textit{i}$.c$ 
            \If{\textit{succ} \textproc{in} OPEN $\land$ \textit{succ}$.c$ < OPEN.\textit{succ}$.c$}
                \State OPEN.\textit{succ} = \textit{succ}
            \ElsIf{\textit{succ} \textproc{in} CLOSE $\land$ \textit{succ}$.c$ $<$ CLOSE.\textit{succ}$.c$}
                \State \textproc{erase}(\textit{succ} \textproc{in} CLOSE) \State \textproc{add}(\textit{succ} \textproc{in} OPEN)
            \ElsIf{$\neg\,$(\textit{succ} \textproc{in} OPEN) $\land$ $\neg\,$(\textit{succ} \textproc{in} CLOSE)}
                \State \textproc{add}(\textit{succ} \textproc{in} OPEN)
            \EndIf
        \EndFor
    \EndFor
    \State \textproc{erase}(\textit{curr} \textproc{in} OPEN)
    \State \textproc{add}(\textit{curr} \textproc{in} CLOSE)
\EndWhile
\State{\textit{opt\_path}.addEdge(getEdge(\textit{curr}))}
\State{\textit{opt\_path}.addNode(\textit{curr})}
\State{SearchState \textit{tmp} = \textit{curr}.father}
\While{$\neg\,$(\textit{tmp} == \textit{start})}
    \State{\textit{opt\_path}.addEdge(getEdge(\textit{tmp}))}
    \State{\textit{opt\_path}.addNode(\textit{tmp})}
    \State {\textit{tmp} $\gets$ \textit{tmp}.father}
\EndWhile
\State{\textit{opt\_path}.addNode(\textit{start})}
\State {return \textit{opt\_path}}
\end{algorithmic}
\end{algorithm}

\alglanguage{pseudocode}
\begin{algorithm} [!t]
\small
\caption{RecursiveAO*}\label{algo:ao*_rec}
\begin{algorithmic}[1]
\Require{\textit{root}, \textit{leaves}}{}
\State Path \textit{opt\_path}
\State \textit{goals} $\gets$ \textit{leaves}
\While {(\textit{goals} $\neq$ \textit{root})}
    \State{\textit{opt\_path} = AO*(\textit{root}, \textit{goals})}
    \State \textit{nextAction} = \textit{opt\_path}.Node.action
    \State \textit{nextAgent} = \textit{opt\_path}.Edge.agent
    \State sendAction(\textit{nextAction}, \textit{nextAgent})
    \State updateGoals(\textit{goals})
\EndWhile
\end{algorithmic}
\end{algorithm}

\begin{figure}[t]
    \centering
    \vspace{2mm}
    \includegraphics[scale=0.55]{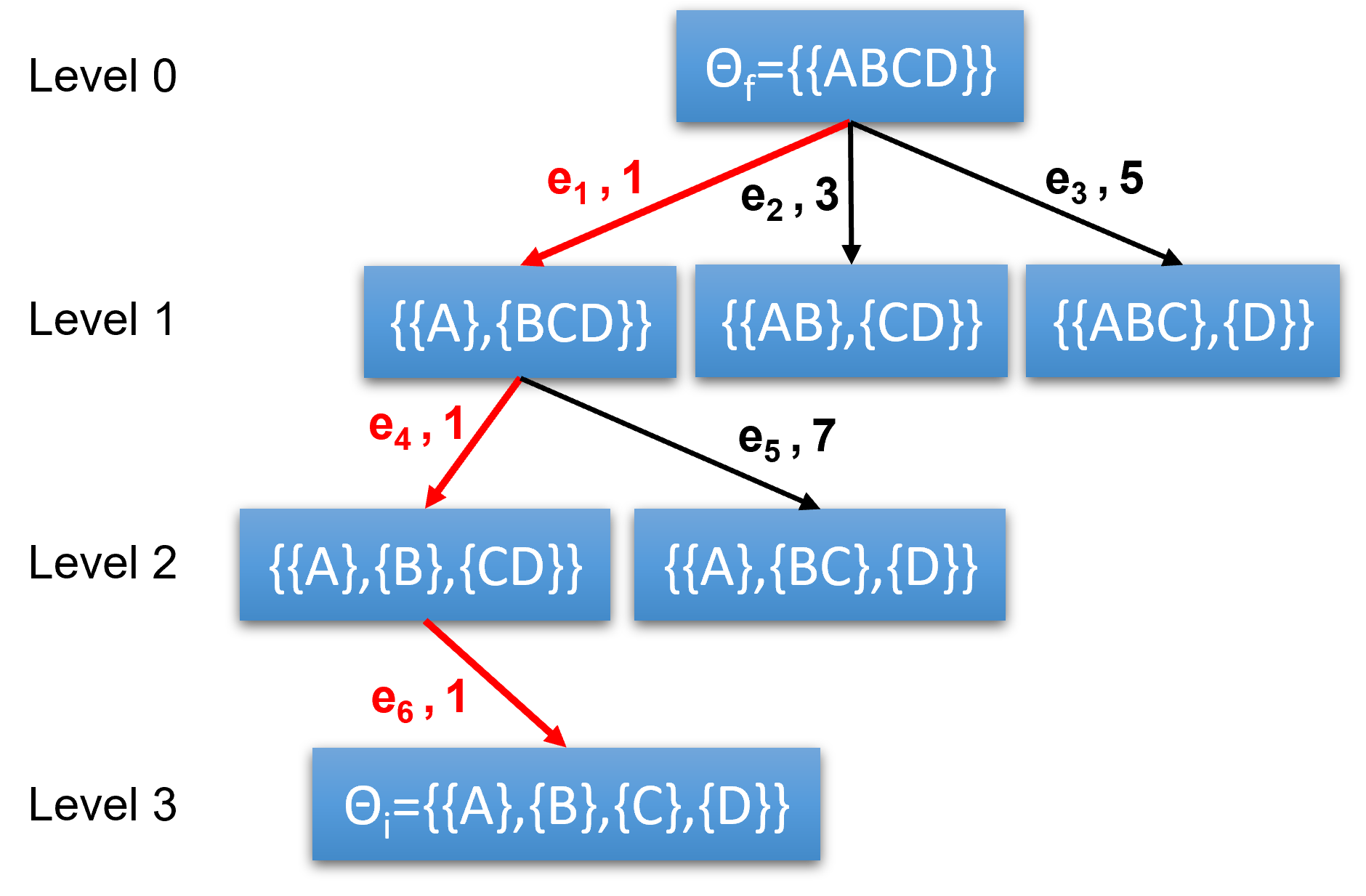}
    \caption{Example of auxiliary tree for a general assembly ABCD, highlighting the optimal path from $\Theta_f$ to $\Theta_i$. The nodes are the ones visited by AO*.}
    \label{fig:ZE}
    \vspace{-4mm}
\end{figure}

The search acts in a top-down fashion, from the \textit{SearchState} (start) containing the assembly configuration $\Theta_f$, which represents the state where all the pieces are assembled, to the \textit{SearchState} (goal) containing $\Theta_i$, the leaf nodes disconnected, that are the inputs of the algorithm.
When a \textit{SearchState} (curr) is visited, the tree is expanded by creating new \textit{SearchStates}. Each of them (succ) contains a feasible configuration $\Theta$ and curr is their father. The cost of succ results from the cost of curr plus the edge cost that connects curr to succ. The new \textit{SearchState} curr to be visited is the one with the minimum score (see Algorithm \ref{algo:ao*}).
Once the \textit{SearchState} goal is reached, the optimal path is obtained by going backwards to the start (\autoref{fig:ZE}). 
The obtained path minimises the sum of the costs of the travelled edges.

While the optimality of the solution, with fixed costs, is ensured, if costs change during the task execution, the search algorithm should be invoked again. In such a case, the algorithm should explore a reduced tree, from the \textit{SearchState} containing $\Theta_f$ to the \textit{SearchState} containing the current assembly configuration (see Algorithm \ref{algo:ao*_rec}). Therefore, each time the AO* search is called, the configuration of the \textit{SearchState} goal $\Theta_i$ contains the current assembly configuration.

\section{Hyper-arcs Cost Computation}
\label{sec:cost_comp}
So far, we have explained how to integrate the role allocation into the planning of the assembly task and how the optimal agent-to-task allocation can be retrieved. However, such a framework is still general and could be exploited to optimise an HRC scenario in different ways. In this context, the design of the cost function plays a crucial role. In particular, in this manuscript, we focus on ergonomics and, hence, we will discuss how to embed ergonomic principles in time-varying hyper-arcs costs. 
Such costs should estimate the ergonomic risk of the human worker at every instant of the collaboration and be able to predict the risk for future actions, and in the case of high risks, delegate such tasks to the robotic agents. Therefore, we propose a continuous risk indicator ${V}(t) \in (0, V_{max})$ for human ergonomic risk and a prediction model using linear difference:
\begin{equation}
 \hat{V}(t_{k+1}) = \alpha_{k+1}{V}(t_{k}) + \beta_{k+1}
 \label{eq:pred_mod}
\end{equation}
where $V(t_k)$ is the risk indicator after achieving action $a_{k}$ (at time $t_k$), $\hat{V}(t_{k+1})$ is the estimated risk indicator after achieving action $a_{k+1}$ (at time $t_{k+1}$), and $\alpha_{k+1}$ and $\beta_{k+1}$ are the linear parameters that regulate the rise of $V(t_k)$ over time, depending only on action $a_{k+1}$. Suitable methods to compute ${V}(t)$ and estimate the prediction parameters are explained in detail in \autoref{sec:erg_assess}. 

Thus, after the execution of action $a_k$, the current value of the ergonomic risk $V(t_k)$ is available and all the $\hat{V}(t_{k+1})$ of all the possible future assembly operations $a_{k+1}$ are computed according to \eqref{eq:pred_mod}. 
Once the predicted risk is known for each possible future action, a rule is needed to update the hyper-arcs cost, accounting for the changing ergonomics. Given any $a_{k+1}$ modelled by the corresponding $h_j$, to execute that $a_{k+1}$ will cost the human worker ($w_h$):
\begin{equation} \label{eq:human_cost}
c_{a_{k+1,w_h}} = c_{h_j,w_h} = 
    \begin{cases}
    \hat{V}(t_{k+1}) \quad &\text{if} \quad \hat{V}(t_{k+1}) <  V_{th} \\
    \hat{V}(t_{k+1}) + \Gamma \quad &\text{if} \quad \hat{V}(t_{k+1}) \geq V_{th}
    \end{cases}
\end{equation}
where $V_{th}$ represents a threshold value, chosen as a certain percentage of $V_{max}$, the maximum value that $V(t)$ can assume, and $\Gamma>V_{max}$. 
The role of $\Gamma$ is that of increasing the hyper-arc cost such that the AO* would be discouraged from selecting that hyper-arc during the search for the optimal path.
As discussed in \autoref{sec:intro}, in the case of mixed human-robot teams, the functions describing agents' suitability for each task must be comparable. 
For simplicity, as robot worker ($w_r$) actions cost, a constant value $c_{h,w_r}$, of the same order of magnitude of $c_{h,w_h}$ is selected, such that $V_{th} < c_{h,w_r} < V_{max} + \Gamma$. The particular choice of its value determines the percentage of robot activity during the collaboration: the lower $c_{h,w_r}$ the higher such percentage, and vice-versa.

\section{Ergonomic Assessment}
\label{sec:erg_assess}
The proposed framework is not coupled with any specific metric; therefore, the choice of the ergonomic index for evaluating the human state during their performance is task-oriented.
As discussed in \autoref{sec:intro}, the description of the human physical state can be kinematic (by considering only body postures), dynamic (evaluating also the effect of the overloading forces), or cognitive (evaluating attention, stress levels, etc.).
For example, in the case of handling, lifting, and transporting heavy objects, assessment methods based on human dynamics are more capable of capturing the ergonomic risk compared to kinematic methods. In this context, muscle~\cite{ma2010new} and joint \cite{lorenzini2019new, lamon2019capability} fatigue models, that depend on muscle activation and overloading torques, represent suitable computational models for $V(t)$. Alternatively, if tasks require a high focus level and expertise, it might be relevant to also monitor physiological stress and workers' attention. For instance, in \cite{lagomarsino2021online} an estimate of cognitive load risk based on gaze direction, head orientation, and body language activities such as self-touching, is proposed. 
Instead, in the case of assembly tasks involving a large number of lightweight pieces, a kinematic index that intrinsically evaluates work frequency can capture potentially damaging situations more precisely. Other potentially relevant factors, such as the weight of the objects to be manipulated, the layout and organisation of the work, are considered less relevant for the risk assessment in the depicted scenario.
Most of the standard risk assessment methods either evaluate the joint configuration only at the current instant, independently of task duration, repetitions, and past activities \cite{mcatamney1993rula, hignett2000rapid},
or adds many factors (action frequency, recovery periods, excerpted forces) which results accurate only if the task execution is well known a priori, e.g. OCRA \cite{occhipinti1998ocra}.
Instead, using the linear prediction previously introduced (1), it is possible to predict the ergonomic risk for future assembly actions, even if the other factors are not known. In light of this, we designed a new risk indicator, KWear (\autoref{ssec:kin_wear}), an index able to combine human posture with action duration. In particular, it focuses on how past activities, as well as repetitions of the same activity, influence the ergonomic assessment of future actions. 

\subsection{Kinematic Wear Index} \label{ssec:kin_wear}
The idea behind the KWear design consists of having an index able to not only describe the current risk by measuring the current body configuration but also convey information about the past posture hazard, and, indirectly, the repetition of sequences of actions, providing a more accurate description than the traditional kinematic indexes such as RULA/REBA \cite{mcatamney1993rula, hignett2000rapid}.
Note that this index does not explicitly address other potentially relevant risk factors, which are neglected in our assessment.
\paragraph{Modelling the Kinematic Wear}
The KWear index at the joint level is modelled as an RC circuit:
\begin{equation}
    \frac{dV_i(t)}{dt} = 
    \begin{cases}
    (1 - V_i(t)) \frac{|G_i(\textbf{q}(t))|}{C} \qquad &\textit{wear}\\
    -V_i(t)\frac{r}{C} \qquad &\textit{recovery}\\
    \end{cases}
\label{eq:kin_wear_diff}
\end{equation}
where $V_i(t) \in [0,1)$, $C$, $r$, and $G_i(\textbf{q}(t))$ are, respectively, the KWear level at instant $t$, the endurance capacity, the recovery rate, and the current risk score of the \textit{i-th} upper body joint assigned by Rapid Upper Limb Assessment (RULA) \cite{mcatamney1993rula}. Such a score is given according to the joint configuration, represented by its rotation about the main Cartesian axes $\textbf{q} = [q_x \hspace{2mm} q_y \hspace{2mm} q_z]$. In particular, the joints evaluated by RULA tables are the shoulder, elbow, wrist, trunk, and neck, and the values $G_i(\textbf{q})$ belong to $[1, 6]$.
In literature, a similar model has been exploited already to describe human muscle activities~\cite{ma2010new} and also thermal motor usage~\cite{urata2008thermal}. 
Therefore, at each time instant $t$, the KWear level is described by:
\begin{equation}
    V_i(t) = 1 - (1-V_i(t_{0}))\hspace{1mm} e^{- \int_0^t{\frac{G_i(\textbf{q}(\tau))}{C}d\tau}}
\label{eq:kin_wear}
\end{equation}
where $V_i(t_0)$ represents the KWear level at the initial instant $t_0$. During action execution, the \textit{i-th} joint motion produces a variation of the RULA score $G_i(\textbf{q}(t))$, which, in turn, increases the corresponding $V_i(t)$, with a slope proportional to the risk level of the new posture: the riskier the posture the steeper the KWear trend. 
Besides, if the human rests, i.e., keeps a comfortable joint configuration, the KWear level of each joint decreases according to the recovery function (RC circuit discharge):
\begin{equation}
    V_i(t) = V_i(t_{0})\hspace{1mm} e^{- \frac{r}{C}t}.
\label{eq:recovery}
\end{equation}
According to \cite{mcatamney1993rula}, in a static configuration, a subject can apply a low force for $T_{max} = 240$ $s$ until feeling physical discomfort. Therefore, the capacity value $C$ is selected to allow each joint wear to approach the asymptotic value $V(T_{max})=1$, starting from $V(0)=0$.
The value $C$ is retrieved by inverting \eqref{eq:kin_wear} with $V_i(t) = 0.993 = V_{max}$ (corresponding to five time constants), and an average level of RULA of $G_{avg} = 3$. Thus, $C$ is the same for all the joints:
\begin{equation}
    C = - G_{avg} \hspace{1mm} \frac{T_{max}}{\ln(1 - V_{max})} = - 3 \hspace{1mm} \frac{240}{\ln(0.007)}.
\label{eq:C_comp}
\end{equation}

For what concerns the recovery rate $r$, it is obtained by inverting \eqref{eq:recovery} considering to match the recovery time (discharge) with the wear time (charge), i.e., the joint can fully recover in a period of $T_{max}$. Therefore, we have:
\begin{equation}
    r = - \frac{C}{T_{max}} \hspace{1mm} \ln \left(\frac{1-V_{max}}{V_{max}}\right),
\label{eq:r_comp}
\end{equation}
The KWear behaviour with the chosen parameters values was validated in experimental tests in~\autoref{ssec:kin_wear_test}. However, they can be optimised to fit subject-specific bio-mechanic characteristics.  

At this point, the KWear prediction model can be obtained by combining \eqref{eq:kin_wear} with \eqref{eq:pred_mod}: 
\begin{equation}
 \hat{V}_i(t_{k+1}) = 1 -\alpha_{k+1,i}(1 - {V}_i(t_{k}))
 \label{eq:kin_wear_pred}
\end{equation}
where $$\alpha_{k+1,i} = e^{-\int_0^{t_{k+1}}{\frac{G_i(\tau)}{C}d\tau}}$$ and $$\beta_{k+1,i} = 1 - e^{- \int_0^{t_{k+1}}{\frac{G_i(\tau)}{C}d\tau}} = 1- \alpha_{k+1,i}$$ are parameters that regulate the raise of KWear over time, for the $i$-th joint. These parameters are representative of the level of the $i$-th joint involvement in the action $a_k$ and hence should be estimated offline with a calibration procedure.

Since the KWear is computed for each monitored joint independently from the others, to recover an overall indicator of the action-related risk, \eqref{eq:human_cost} becomes:
\begin{equation} \label{eq:human_cost_KWear}
c_{a_{k+1,w_h}} = c_{h_j,w_h} = \sum_{i=1}^{m} \gamma_{i},
\end{equation}
where $m$ is the number of monitored joints and
\begin{equation} \label{eq:kin_wear_ranges}
    \gamma_i =
    \begin{cases}
    \hat{V}_i(t_{k+1}) \quad &\text{if} \quad \hat{V}_{i}(t_{k+1}) <  V_{th,i} \\
    \hat{V}_i(t_{k+1}) + \Gamma \quad &\text{if} \quad \hat{V}_{i}(t_{k+1}) \geq V_{th,i}.
    \end{cases}
\end{equation}

\paragraph{Calibration Procedure}
\hspace{3mm} The calibration procedure is needed to estimate the parameters of the KWear prediction model $\alpha_{k,i}$ (and consequently $\beta_{k,i}$) for each action $a_k$ and each joint $i$ which depend only on the $k$-$th$ task.
The idea is to record offline the joint angles (and the RULA scores) in $\eta$ executions of the same assembly action, compute $\alpha_{k,i}$ and average them to obtain a value able to approximate a nominal execution of the task.
This implies that our method could not adapt online if workers change how they solve a task, e.g., because of increasing tiredness. In this work, we assume that the different executions of tasks are demonstrated in the calibration procedure.
To design the number of executions $\eta$ necessary to ensure at maximum the prediction error $\Tilde{V}_{i,des}(t_{k+1})$, it is possible to adhere to the following strategy. 
First, the target maximum prediction error $\Tilde{V}_{i,des}(t_{k+1})$ is fixed. This value represents the level of accuracy we expect from the system when estimating the KWear during the risk prediction phase. 
Second, the action is executed $\eta$=$\eta_0$ times at null initial conditions, where $\eta_0 \geq 2$ represents the minimum number of repetitions that includes possible different executions of the action. 
Then, the mean $\alpha_{k,i,\eta}$ is plugged in model \eqref{eq:kin_wear_pred} to obtain $\hat{V}_{i,\eta}(t_{k+1})$. Such a value can be then compared with the actual $V_{i,l}(t_{k+1})$ values recorded for all the $\eta$ executions, where $l \in [1, \eta]$, resulting in the prediction errors $\Tilde{V}_{i,l,\eta}(t_{k+1})$. If $\Tilde{V}_{i,l,\eta}(t_{k+1})$ $>$ $\Tilde{V}_{i,des}(t_{k+1})$ at least for one value of $l$, the computations should be performed again with another task execution ($\eta$=$\eta$+$1$), until each $\Tilde{V}_{i,l,\eta}(t_{k+1})$ $<$ $\Tilde{V}_{i,des}(t_{k+1})$ or a maximum $\eta_{max}$ is reached.

Please note that the calibration phase has to be repeated when a new worker is involved in the collaboration. This is due to the fact that the prediction parameters must be descriptive of how each specific action is performed by the current worker.
For a more accurate risk prediction in real-world applications where more than one worker is assigned to the same task, it is convenient to use person-specific model parameters. A possible solution consists of creating a database containing all the subject-parameters correspondences and then selecting the correct one once the worker changes.
Alternatively, the parameters could be estimated during the online task performance and updated through a weighted average to prioritise the $\alpha_{k,i}$ computed during the last execution of action $a_k$. In this way, the system would learn how each worker is used to execute each action.

\section{Computational Complexity Evaluation}
\label{sec:comp_compl}
The computational complexity of our role allocation algorithm was tested to evaluate the performances of the proposed method. We carried out the simulations on a desktop with an Intel Xeon W-2245 3.9 GHz x 16-cores and 64 GB of RAM. The software architecture is based on ROS Melodic running on Ubuntu 18.04. 

The main computational blocks of our algorithm are three: the ergonomic assessment, returning the KWear level at each monitored joint; the hyper-arcs cost computation, which updates actions cost of the corresponding hyper-arcs; and the AO*, returning the instruction for the next action.
In particular, the ergonomic assessment process and the hyper-arcs cost updating process were implemented as nodes in Python, while the AOG library and AO* algorithm were written as a C++ node, to speed up the heavy computations required by the algorithm. 

Unlike the first two processes, that are continuously executed at high frequency ($20$ Hz) during the whole task duration, the AO* search is called only when action $a_k$ is completed, with hyper-arcs cost updated with the current human ergonomic state.
For this reason, to evaluate the computational complexity of the framework, we will focus on the performance of the AO*, tested in different conditions.

To provide a comprehensive description two different task representations are evaluated, which differ on how pieces may be interconnected~\cite{demello1991correct}:
\begin{itemize}
    \item One case is an assembly with \textit{sequentially} connected atomic pieces in which there are N-1 interconnections between the N pieces, with the i-th interconnection connecting part $p_i$ and $p_{i+1}$~\cite{demello1990and}. An example of this task model is represented by the assembly of a pen, where each of the 4 atomic pieces (cap, ink, body, and bottom) can be assembled equivalently with the piece before or after within the sequence. 
    \item The other case is an assembly with \textit{scarcely} connected atomic pieces, where all the N pieces are not directly connected among themselves but only to a specific sub-assembly. An example of this second category of assembly is a table (top and 4 legs), where the legs cannot be assembled together but only with the top.
\end{itemize}
Moreover, we tested the search by increasing:
\begin{enumerate}[i.]
    \item \textit{leaf nodes}, from 2 to 20, representing the atomic assembly pieces (with a fixed number of agents equal to 2); 
    \item \textit{agents} involved in the cooperation, from 2 to 30 (with a fixed number of assembly pieces equal to 10). 
\end{enumerate}

\begin{figure}[t]
    \centering
    \includegraphics[trim=0.1cm 0.0cm 0.7cm 0.7cm,clip,width=
    \columnwidth,height=5.5cm]{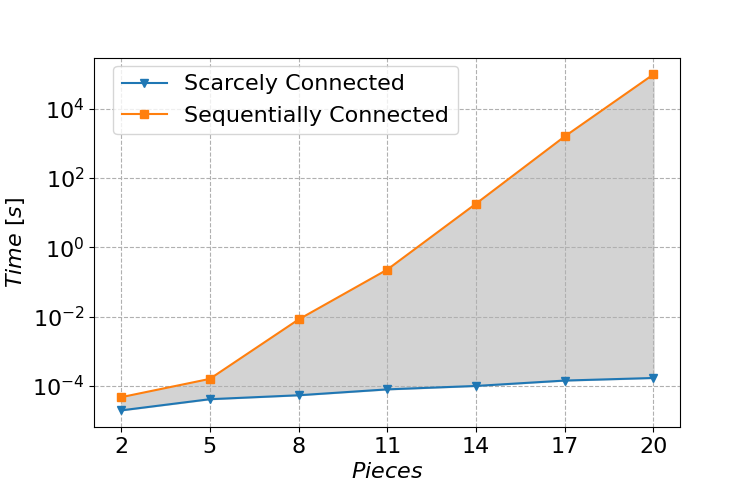}
    \includegraphics[trim=0.1cm 0.0cm 0.7cm 0.7cm,clip,width=
    \columnwidth,height=5.5cm]{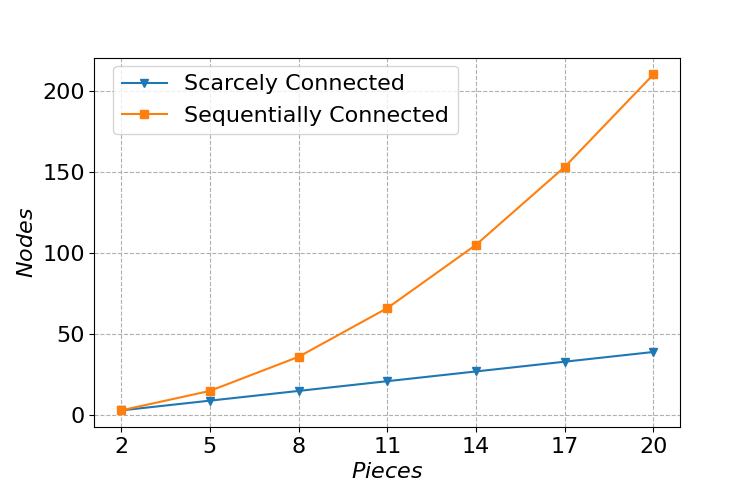}
    \includegraphics[trim=0.0cm 0.0cm 0.7cm 0.7cm,clip,width=
    \columnwidth,height=5.5cm]{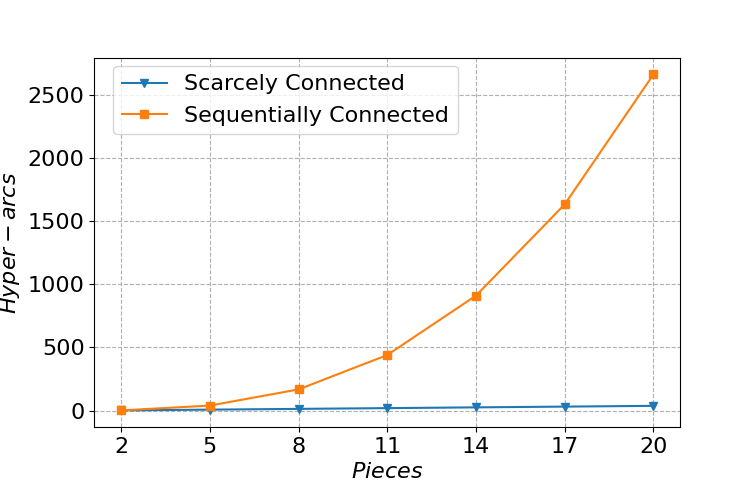}
    \caption{(top) AO* computational time in log scale obtained by increasing the number of pieces to be assembled in the two cases of sequentially and scarcely connected assembly. (middle) Nodes complexity. (bottom) Hyper-arcs complexity.}
    \label{fig:computational_time}
    \vspace{-2mm}
\end{figure}
\begin{figure}[t]
    \centering
    \includegraphics[trim=0.1cm 0.0cm 0.7cm 0.7cm,clip,width=
    \columnwidth,height=5.5cm]{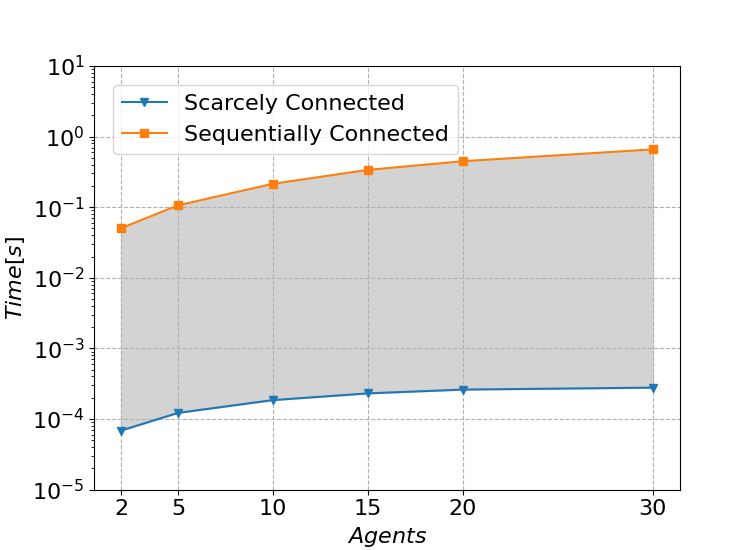}
    \caption{AO* computational time in log scale obtained by increasing the number of agents in the two cases of sequentially and scarcely connected assembly.}
    \label{fig:computational_time_ag}
    \vspace{-2mm}
\end{figure}

The results in \autoref{fig:computational_time} (top) show different curves. In case of sequentially connected assemblies, the trend is exponential.
In the scarcely connected case, instead, the trend is flatter, with a log-linear complexity.
This is because, in the latter case, as the number of pieces $M$ increases, the nodes and the hyper-arcs increase linearly, instead of exponentially as in the first case (see \autoref{fig:computational_time} (bottom)). A in-depth analysis of the completeness and complexity of the plan generation can be found in~\cite{demello1991correct}.
On the other hand, the plot in \autoref{fig:computational_time_ag} presents, in both cases, a linear trend.
This is due to the fact that an increase of number of pieces causes an increase of both hyper-arcs and nodes, while an increase of the number of agents entails only a raise of the number of hyper-arcs.
An assembly with 20 pieces and 2 agents corresponds, in the sequentially connected case, to an AOG with more than 200 nodes and more than 2500 hyper-arcs. Therefore, the optimal search on such a graph is a considerably large problem. On the other hand, if the assembly belongs to the scarcely connected category, with the same number of atomic components, nodes become 39 and hyper-arcs 38, and the AO* presents comparable results as in the case of fewer pieces. 
It can be noticed that a general assembly, made of scarcely and sequentially connected parts, will present results in the grey area between the two curves. This means that, for assemblies of a large number of pieces ($>20$), it is reasonable to reduce the number of feasible assemblies to ensure that the AO* provides results at the proper time. For instance, if an action implies a high risk for the human workers, the related hyper-arcs can be pruned from the AOG, ensuring that the action will be allocated to the robot.
Additionally, it should be noted that the search algorithm acts on an AOG with a smaller size after an action is completed; as a result, the time to generate the solution iteratively is rapidly decreasing.

\section{Experimental Evaluation}
\label{sec:expe}
The proposed framework was evaluated with two distinct experiments in a collaborative proof-of-concept task: (i) validation of the KWear, and (ii) the assessment of the ergonomic role allocation.
In both cases, the collaborative task was the assembly of a corner joint ($J$) with three aluminium profiles, i.e. two sides ($S_1$, $S_2$) and a leg ($L$). The profiles $S_1$ and $S_2$ have the same length, while $L$ is shorter. Each of them is interlocked in a predefined hollow of the corner joint $J$ as shown in \autoref{fig:setup2} (right).  
The online ergonomic assessment was achieved by measuring the configurations of the human main upper-body joints (trunk, neck, and shoulder, elbow, and wrist of the dominant arm) through an inertial-based motion-capture system (the Xsens suit).

\subsection{Kinematic Wear Validation}
\label{ssec:kin_wear_test}
The purpose of this experiment is to demonstrate the KWear potential to recognise ergonomically hazardous situations based on the history of joint configurations.
A single right-handed subject was asked to perform repetitively the aforementioned assembly task in a human-robot cooperative scenario. The agents roles were assigned by the proposed framework, with prediction parameters previously estimated during a calibration phase.

\begin{figure}
\centering
	\includegraphics[trim=0.0cm 0.0cm 0.0cm 0.0cm,clip,width=0.60\linewidth]{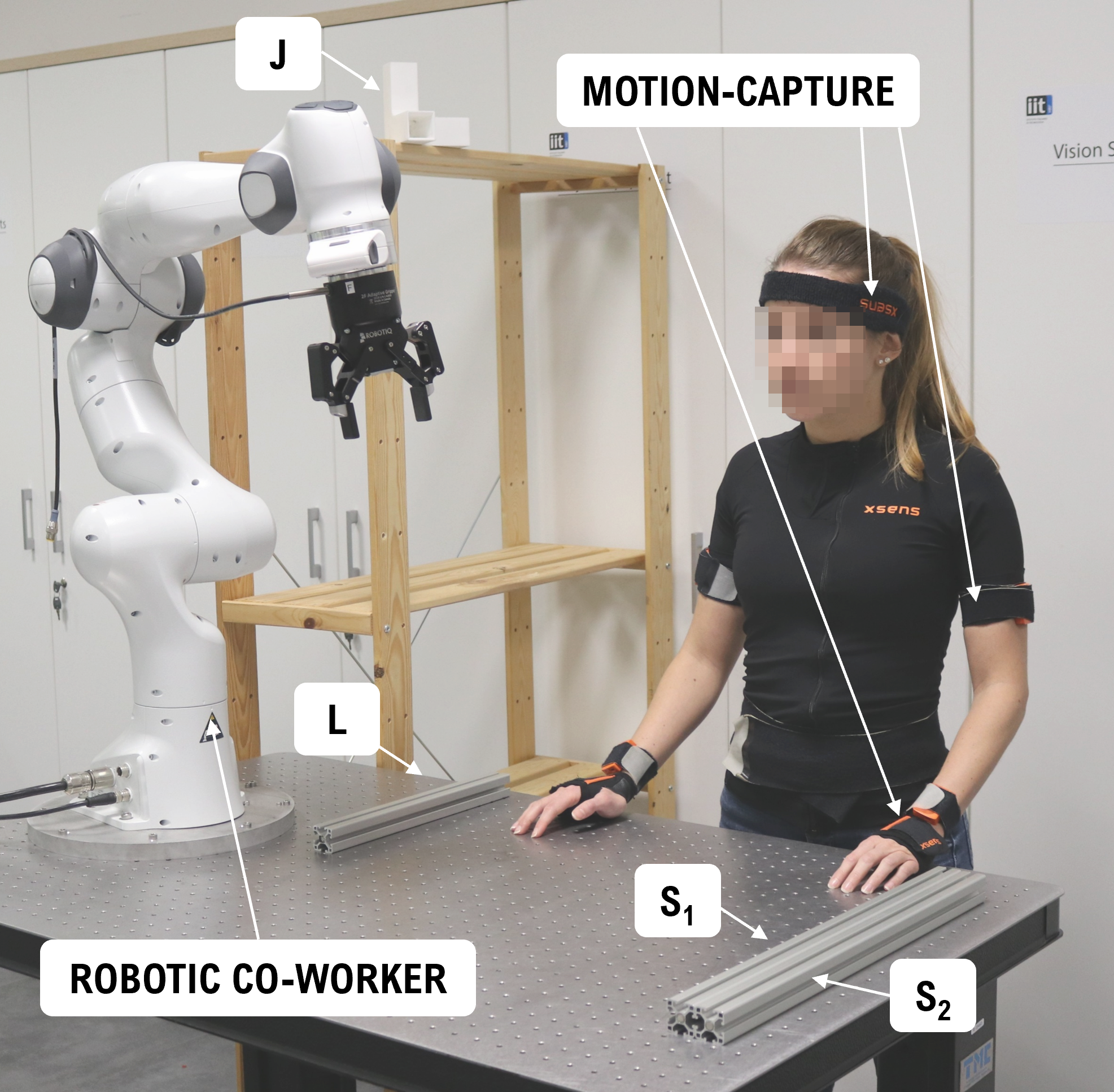}
	\includegraphics[scale=0.2]{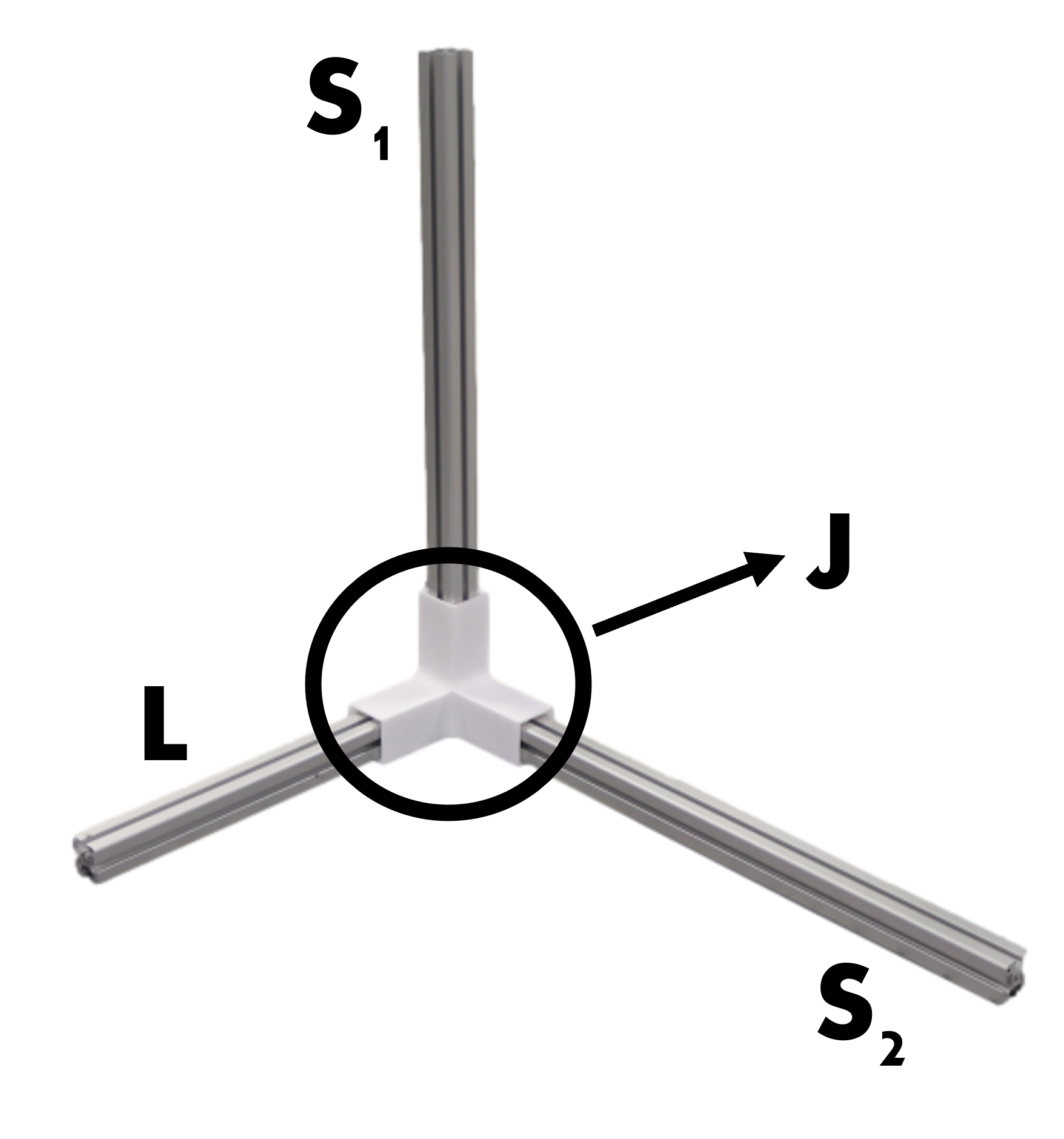}
	\caption{(left) Experimental setup of the KWear Validation experiment. The workstation consisted of a workbench in front of the human operator, shared with the robotic co-worker, and a shelf to the human's right. The subject wore a motion-capture system to record the joint angles. (right) The complete assembly.}
    \label{fig:setup2}
    \vspace{-3mm}
\end{figure}

\paragraph{Experimental Setup and Protocol}
The experimental setup is shown in \autoref{fig:setup2} (left).
The subject was asked to wear the Xsens suit and perform 4 times ($rp_1$, $rp_2$, $rp_3$, and $rp_4$) the assembly task. As a robot co-worker, the Franka Emika Panda manipulator was selected, equipped with a Robotik two-finger gripper.
The fixed sequence of atomic actions of the task were: ($a_1$) pick the corner joint $J$ from its initial position and place it on the workbench closer to the human operator; ($a_2$) pick and insert $L$ in the hollow of $J$ at the operator's right; ($a_3$) pick and insert $S_1$ in the upper hollow of $J$; ($a_4$) pick and assemble $S_2$ in the hollow of $J$ at the operator's left; ($a_5$) pick the complete assembly and place it on another ledge of the shelf. 
The rationale behind this specific choice of actions and object positioning is the achievement of risky posture by imposing some joints close to the range-of-motion (RoM) limit, i.e. with a high level of risk according to RULA. The calibration procedure in~\autoref{sec:erg_assess} was performed to estimate the KWear prediction model parameters.
The hyper-arcs costs were assigned as described in \autoref{sec:erg_assess}, with $\Gamma = 100$, $V_{th,i} = 0.8$ $\forall i$, chosen as the $80\%$ of the maximum value of the KWear, and $c_{h,w_r} = 50$.
Finally, the allocation results are compared to the ones obtained through a RULA-based role allocation, where the ergonomic risk related to the $k$-th action, $G(a_k)$, is computed offline according to the standard method ~\cite{mcatamney1993rula}; the scores assigned to $a_1$, $\dots$, $a_5$ are in Table 1(top). In this case, actions are assigned to the robot when the action-related score is higher than the $80\%$ of the maximum RULA score ($G_{max}$=$9$), that is $G_{th}$=$7.2$, analogously to the KWear threshold.

\begin{table}[!t]
    \begin{center}
    \begin{tabular}{|c|c||c|c|c|c|c|}
    \multicolumn{2}{c}{} & \multicolumn{5}{c}{Ergonomic Risk}\\
    \hline
     \multicolumn{2}{|c||}{} & $\boldsymbol{a_1}$ & $\boldsymbol{a_2}$ & $\boldsymbol{a_3}$ & $\boldsymbol{a_4}$ & $\boldsymbol{a_5}$\\
    \hline
        \multicolumn{2}{|c||}{RULA} & 5 & 3 & 3 & 4 & 4\\
    \hline 
    \multicolumn{2}{c}{} & \multicolumn{5}{c}{Allocation Results}\\
    \hline
    \multirow{4}{*}{KWear}
    &    \textbf{$rp_1$} & \cellcolor{green!50}H & \cellcolor{green!50}H & \cellcolor{green!50}H & \cellcolor{green!50}H & \cellcolor{green!50}H\\
    &    \textbf{$rp_2$} & \cellcolor{red!50}R & \cellcolor{green!50}H & \cellcolor{green!50}H & \cellcolor{green!50}H & \cellcolor{red!50}R\\
    &    \textbf{$rp_3$} & \cellcolor{green!50}H & \cellcolor{green!50}H & \cellcolor{green!50}H & \cellcolor{red!50}R & \cellcolor{green!50}H\\
    &    \textbf{$rp_4$} & \cellcolor{green!50}H & \cellcolor{red!50}R & \cellcolor{green!50}H & \cellcolor{green!50}H & \cellcolor{green!50}H\\
    \hline
        \multicolumn{2}{|c||}{RULA} & \cellcolor{green!50}H & \cellcolor{green!50}H & \cellcolor{green!50}H & \cellcolor{green!50}H & \cellcolor{green!50}H\\
    \hline
    \end{tabular}
    \end{center}
    \caption{(top) Ergonomic risk score of each action computed according to RULA tables. (bottom) Comparison between the allocation results according to the RULA scores and the KWear values. Since KWear can consider the risk related to the past configurations, the allocation solutions change through the repetitions $rp_i$.}
    \label{table:allocation_res}
    \vspace{-4 mm}
\end{table}

\begin{figure*}[t]
\centering
	\includegraphics[scale=0.32]{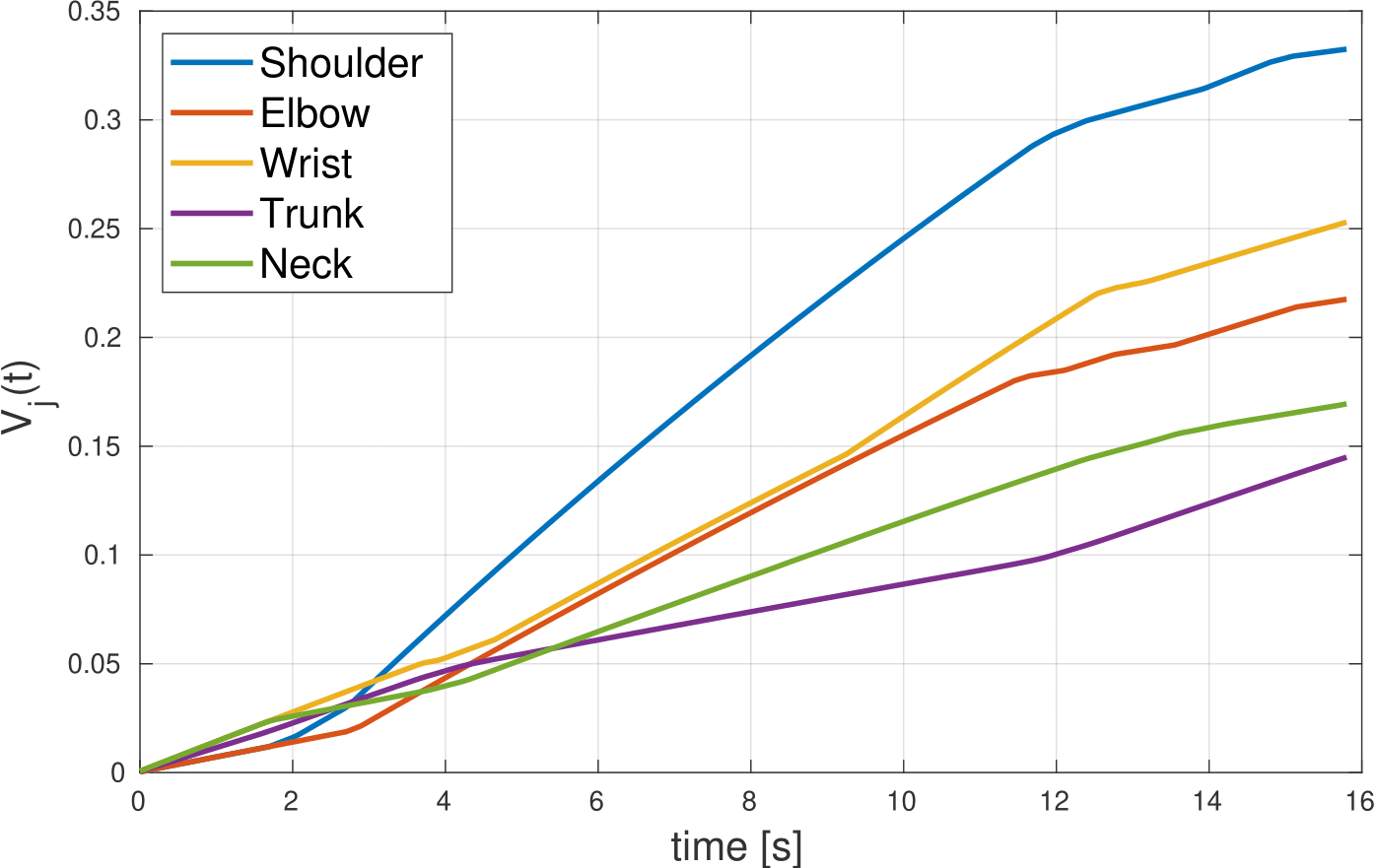}\includegraphics[scale=0.32]{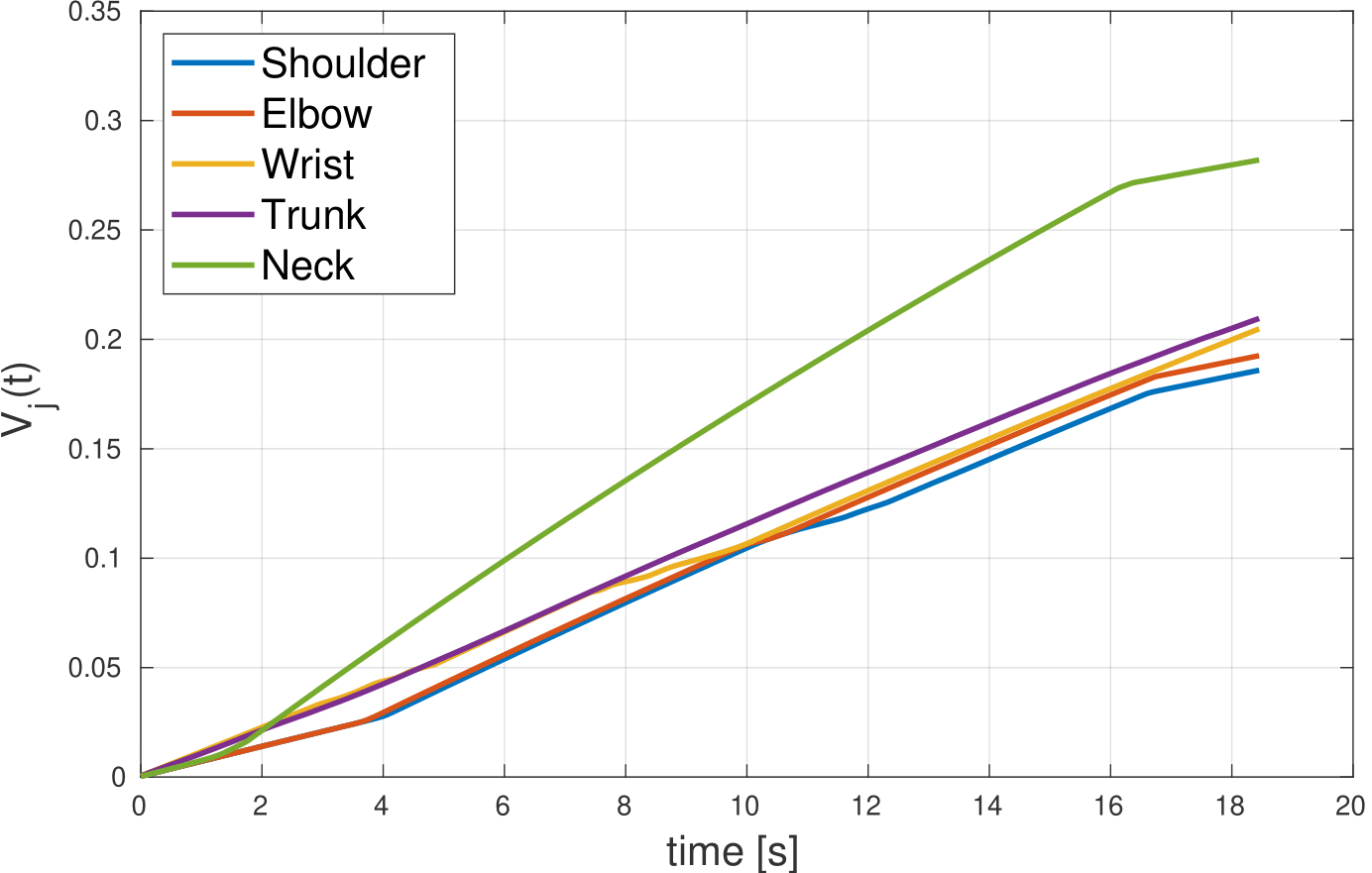}\includegraphics[scale=0.325]{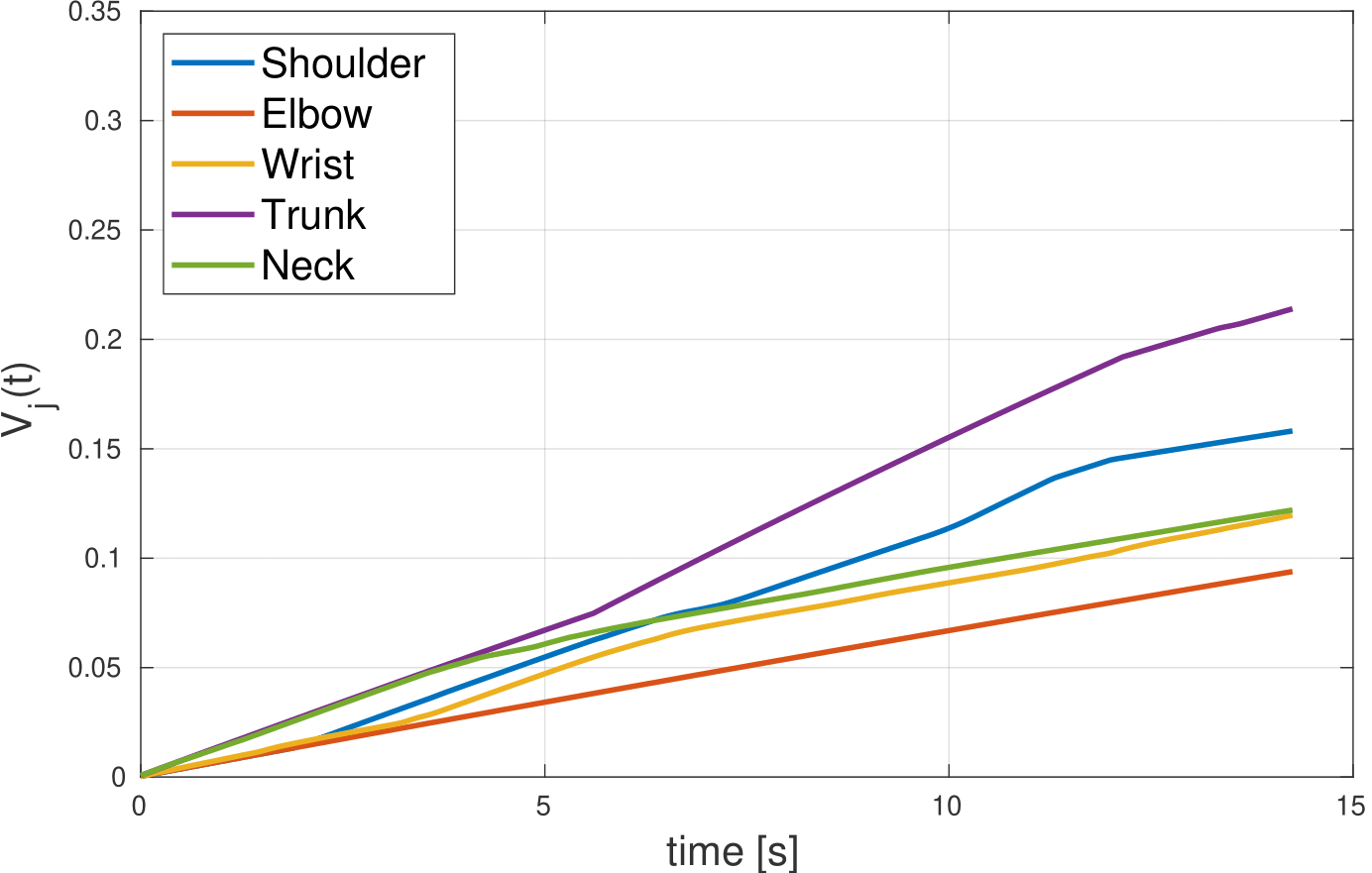}\\
	 \hspace{5 mm}\includegraphics[scale=0.032]{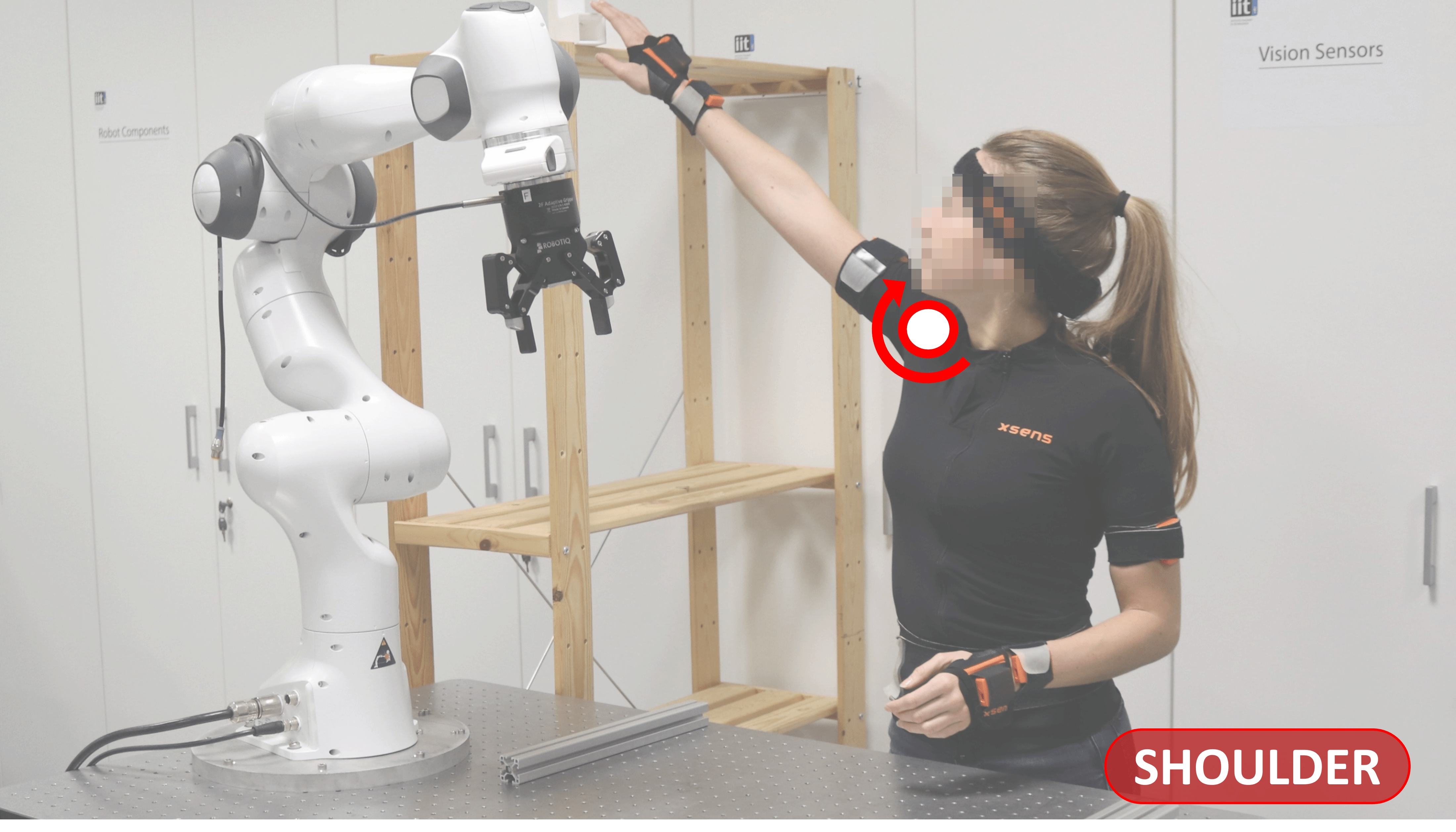} \hspace{5 mm}\includegraphics[scale=0.032]{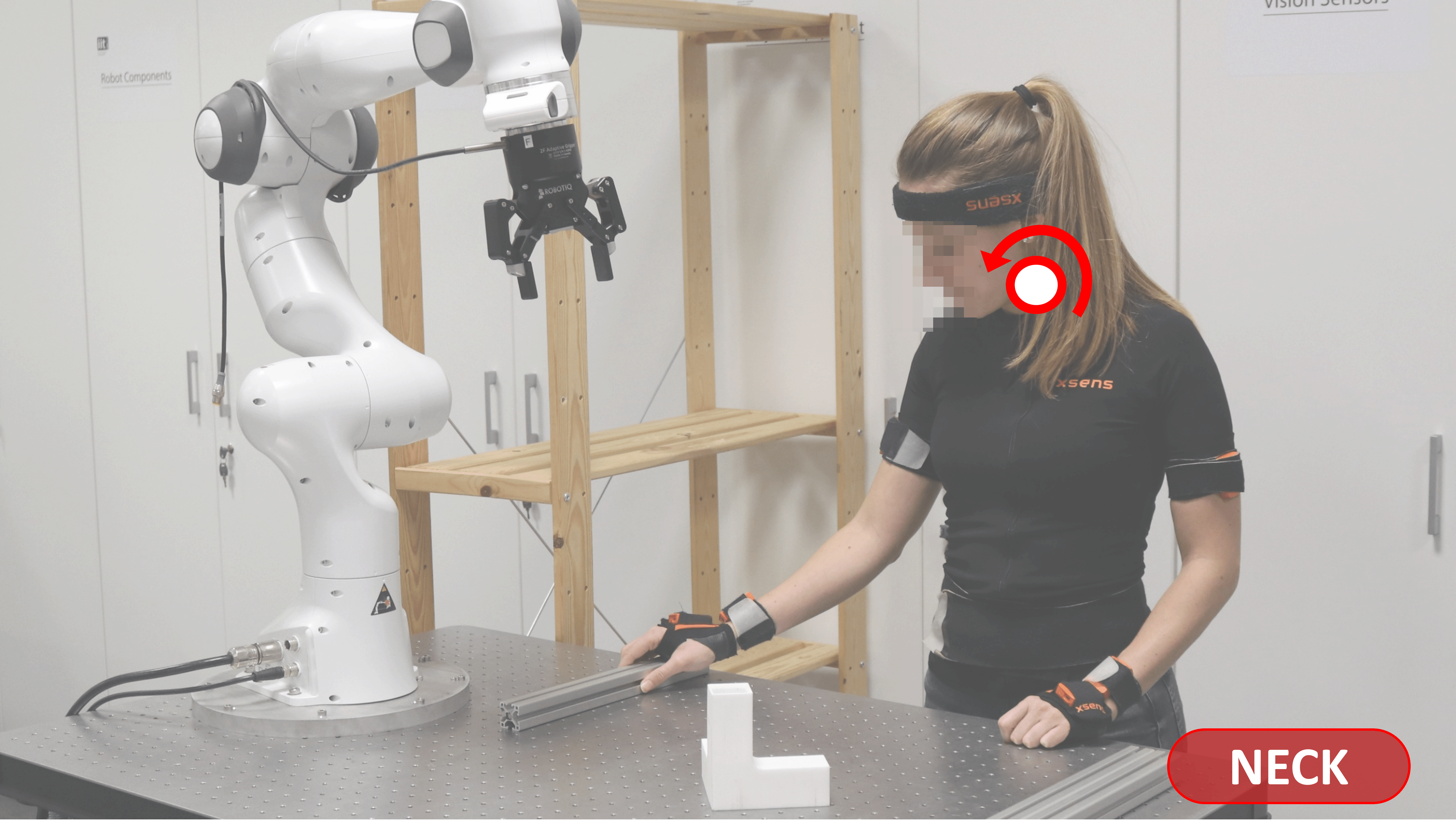} \hspace{5 mm}\includegraphics[scale=0.032]{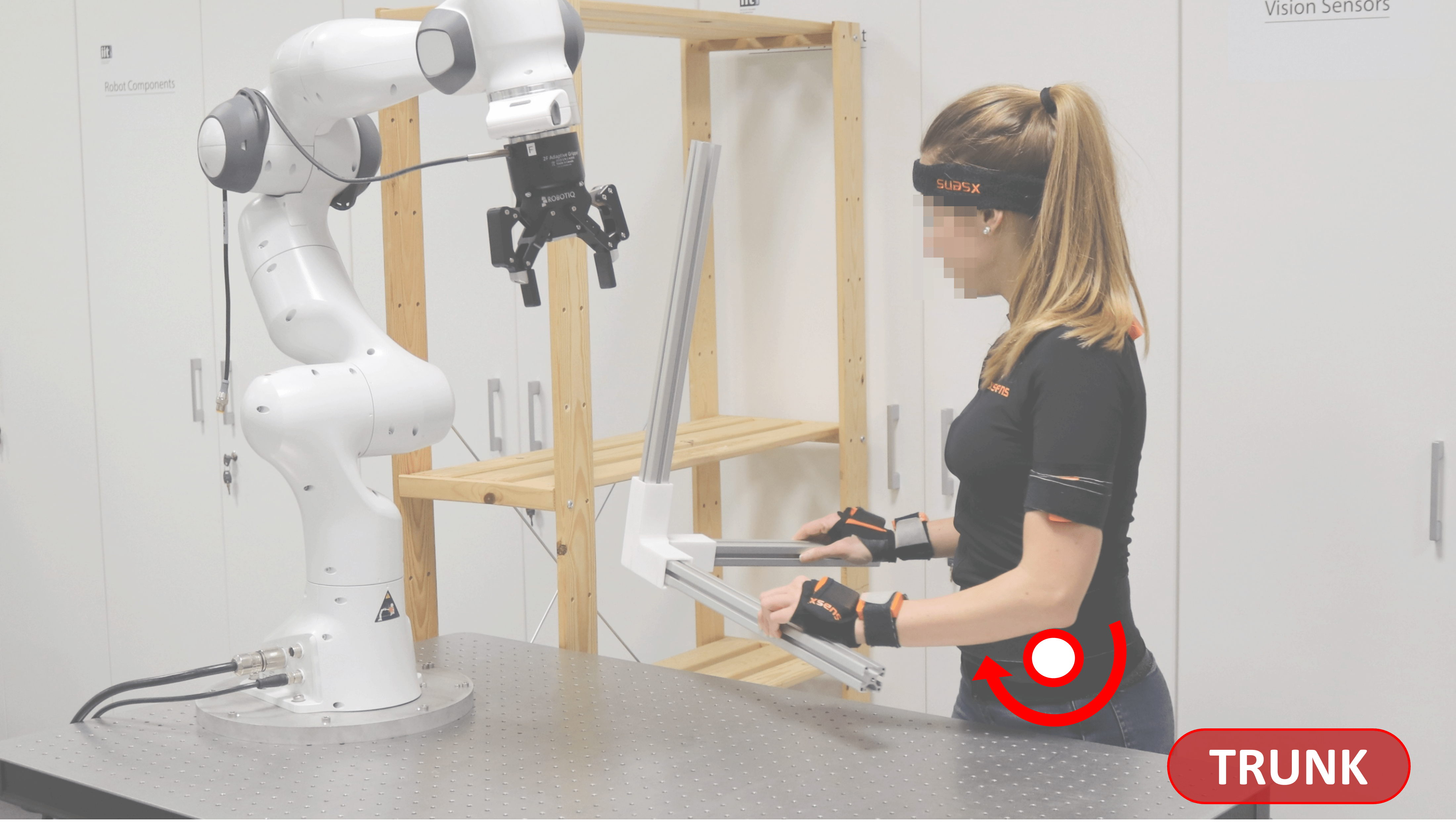}\\
	\caption{The KWear trend for all the monitored joints during the performance of $a_1$, $a_2$, $a_5$, starting from zero initial conditions (top); a frame of the correspondent action showing the higher involvement of a specific joint (bottom).}
    \label{fig:KWear125}
\end{figure*}

\paragraph{Results}
By changing the steepness of the slope, the KWear can describe different levels of risk according to the RULA scores. In particular, in \autoref{fig:KWear125} the KWear trend is shown for the task actions $a_1$, $a_2$ and $a_5$.
The first action ($a_1$) required the subject to raise her arm to reach object $J$ on top of the shelf, therefore the shoulder assumed the riskiest configuration according to RULA tables.
The second action ($a_2$) reported a greater involvement of the neck since the subject had to bend it to localise and reach object $L$, closer to her. The last action ($a_5$) involved mostly the trunk since the human was required to twist it, reaching a RULA region with high risk.

In \autoref{table:allocation_res}, we present the results in terms of agents allocation during HRC for the 4 repetitions of the task. During $rp_1$, the ergonomic risk remained below the threshold for all joints, thus the whole task was allocated to the worker. Conversely, starting from $rp_2$, the risk level increased and some actions began to be allocated to the robot. Notably, when an action was allocated to the robot, the subject could rest, and this recovery produced a decrease in KWear, reducing the ergonomic risk baseline for future actions. 

Regarding the allocation results obtained using a RULA-based role allocation, since all the five actions present a constant RULA value below the threshold, the whole task is assigned to the human worker for each repetition.  

\paragraph{Discussion}
The results suggest that (i) the riskier the joint angle values, the steeper the KWear increment, and (ii) the KWear value rises as the duration of the action increases.
Hence, the model can be used to describe joint usage over time. For instance, $a_1$ stressed the shoulder, $a_2$ the neck, and $a_5$ the trunk (\autoref{fig:KWear125}). 
Although the design of such an index is inspired by the charge-discharge behaviour of muscle fatigue, the carried information is radically different from these biomechanical models.
However, the results highlight the potential of the proposed indicator in evaluating instantaneous ergonomic risk and in predicting future ones, intending to prevent possible worker harm.

\subsection{Multi-Subject Assessment}
The goal of this experiment is to test the complete framework through an extensive multi-subject human study.
12 healthy subjects (2 females and 10 males, $28 \pm 3$ years, right-handed) were asked to execute the assembly both by themselves (\textit{H only}) and in cooperation with the robot (\textit{HRC}).
While 10 subjects had already experienced human-robot interaction applications, all of the subjects were naive with respect to the goal of the experiment.
All experiments were performed at the Human-Robot Interfaces and Physical Interaction (HRII) laboratory, Istituto Italiano di Tecnologia (IIT), and the protocol was approved by the ethics committee of Azienda Sanitaria Locale (ASL) Genovese N.3 (Protocol IIT\_HRII\_ERGOLEAN 156/2020).

\begin{figure}
\centering
	\includegraphics[scale=0.07]{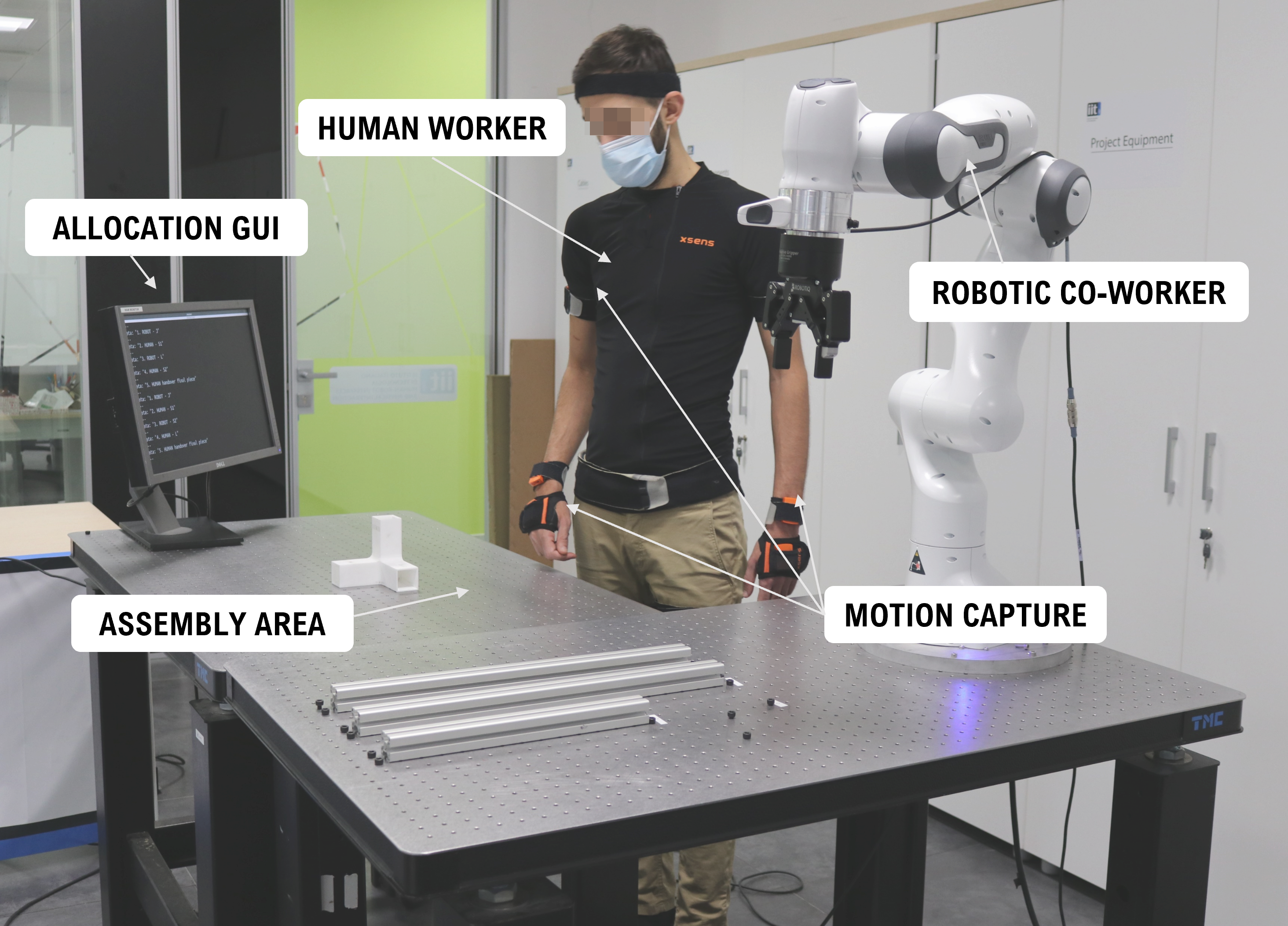}
	\caption{Experimental Setup. The assembly pieces were placed on the operator's left, as well as the robot co-worker; in front of the human, the assembly area and the monitor to read the allocations were located. The human wore a motion-capture system to record the joint angles.}
	\label{fig:exp_setup_3}
	\vspace{-2mm}
\end{figure}

\paragraph{Experimental Setup and Protocol}
The experimental setup is shown in
\autoref{fig:exp_setup_3}. 
The subjects wore the Xsens suit and executed the assembly 10 times for each condition, \textit{H only} and \textit{HRC} in collaboration with the Franka Emika Panda manipulator. To avoid biases, half of the subjects started with the \textit{HRC} modality, the other half with the \textit{H only} one.
The five actions necessary to conclude a repetition were: ($a_1$) pick and place $J$ in the assembly area; ($a_2$) pick and insert $L$ in the upper hollow; ($a_3$) pick and insert $S_1$ in the left hollow; ($a_4$) pick and insert $S_2$ in the right hollow; ($a_5$) move away the entire assembly. In particular, the last action envisioned the handover of the entire assembly to another human operator, who was in charge of disassembling the final product and repositioning the pieces at their original sites to simulate an industrial conveyor roller.
The positions of the assembly pieces to be handled by the robot and the handover position (chosen to make the handover as natural as possible) were pre-planned.
In terms of ergonomic risk evaluated through RULA tables, action $a_2$, since it requires grasping the object positioned in the furthest location from the human, and inserting it in the upper hollow of $J$, exposes the shoulder to a higher risk for its whole duration ($G(a_2) = 5$), while actions $a_3$ and $a_4$ required to handle objects close to the worker ($G(a_3) = G(a_4) = 3$), therefore they are less risky. Actions $a_1$ and $a_5$ have scores $G(a_1) = G(a_5) = 4$.
Before the experiment, the calibration procedure explained in~\autoref{sec:erg_assess} was performed. The subjects were asked to repeat each action of the task $\eta$=$3$ times, which ensured $\Tilde{V}_{i,des}(t_{k+1})$ $<$ $10^{-3}$, selected according to the explained strategy with $\eta_0$=$2$.
To replicate the time constraints of production lines, typical of industrial settings, we introduced the takt time, i.e., the total time in which the assembly has to be completed according to the process demand. In this experiment, we set $T_{takt}$=$38 s$ according to the average time the robot took to execute the whole assembly with a conservative speed of $0.05$ $m/s$. 
 
Unlike the previous experiment, the sequence of actions was not fixed. Only actions $a_1$ and $a_5$ were fixed as the first and last of the assembly sequence, while the order of the other three actions was decided online by the AO*. Each action could be assigned either to the human or the robot, except for the last one, which is always assigned to the human.
The hyper-arcs costs were updated as described in \autoref{sec:erg_assess}, while the cost of robot hyper-arcs was tuned to balance the human and robot activity within the collaboration, in particular, $c_{h,w_r}$=$2.5$. 

\begin{figure}[t]
\centering
\includegraphics[trim=0.0cm 0.0cm 0.0cm 0.1cm,clip,width=0.49\columnwidth]{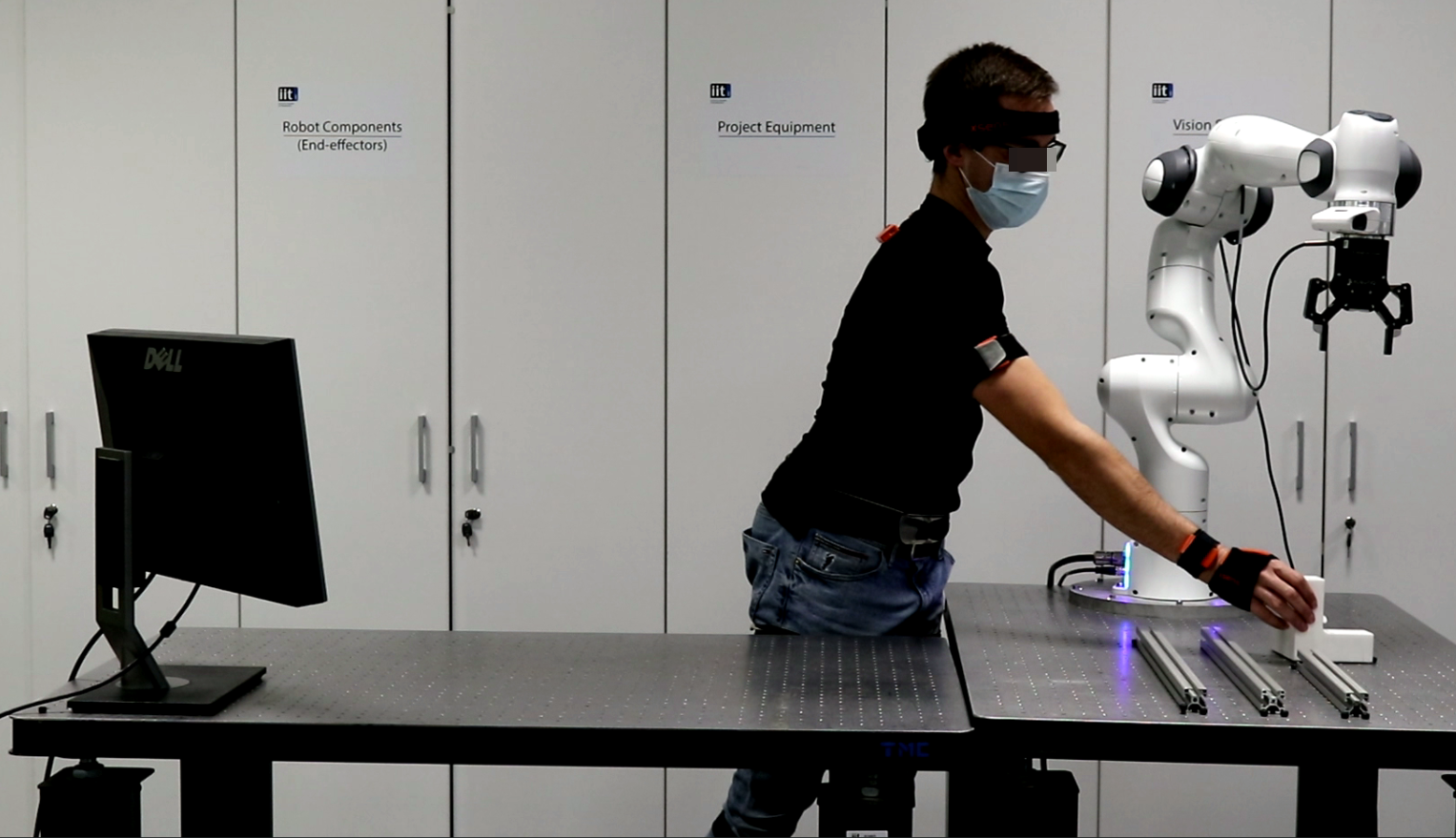}
\includegraphics[trim=0.0cm 0.0cm 0.0cm 0.0cm,clip,width=0.49\columnwidth]{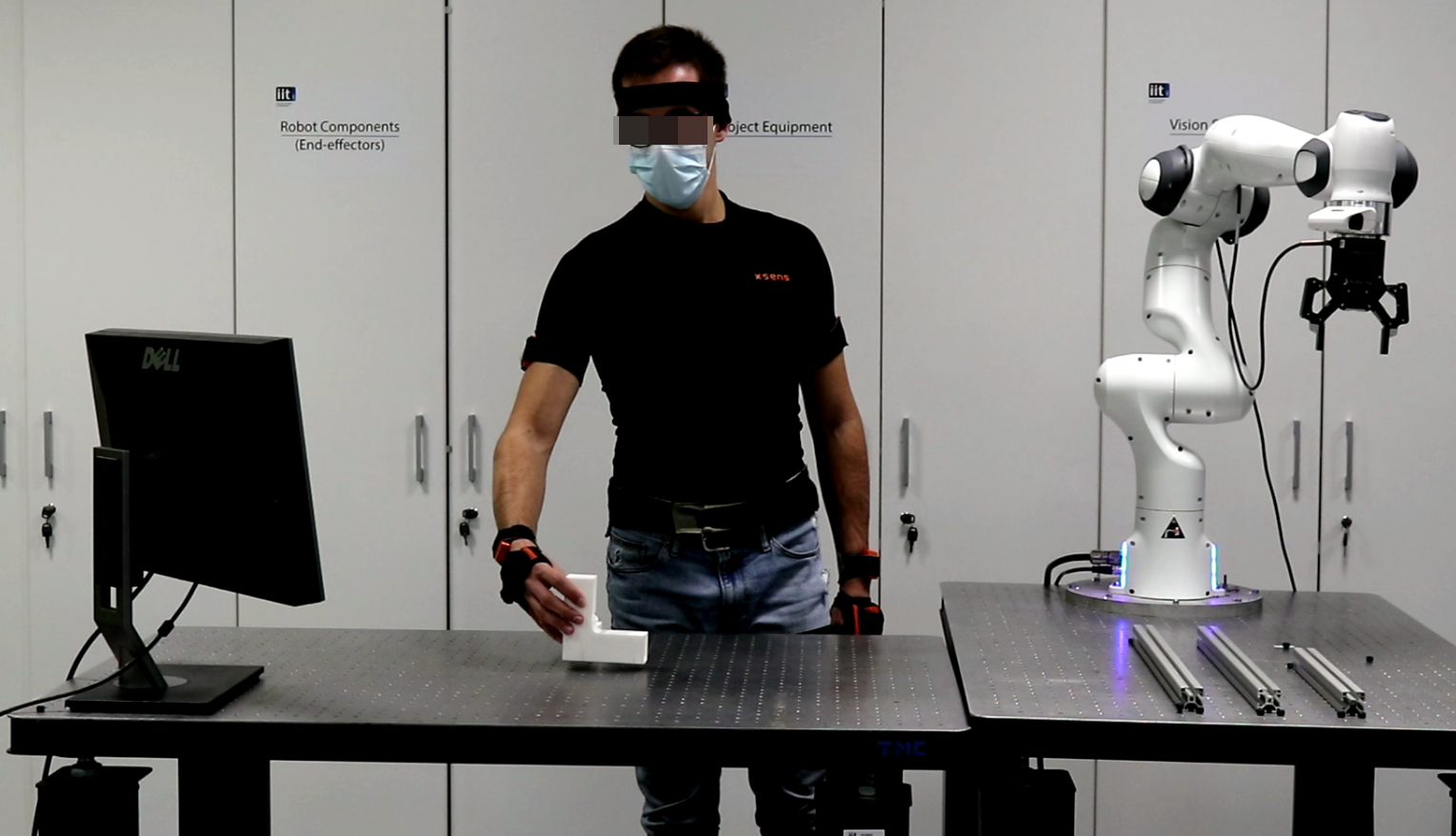}
\\
\includegraphics[trim=0.0cm 0.0cm 0.0cm 0.0cm,clip,width=0.49\columnwidth]{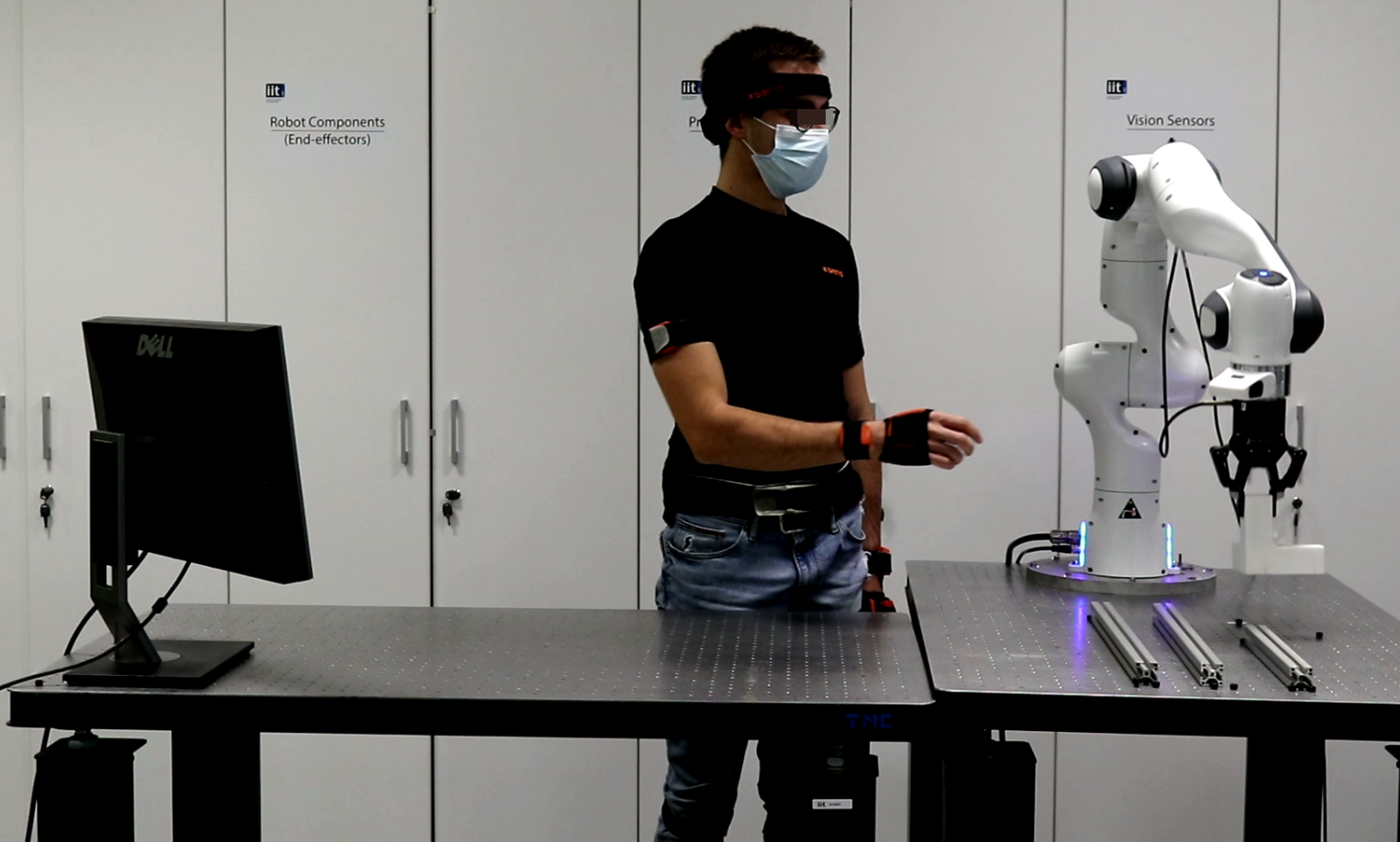}
\includegraphics[trim=0.0cm 0.0cm 0.0cm 0.0cm,clip,width=0.49\columnwidth]{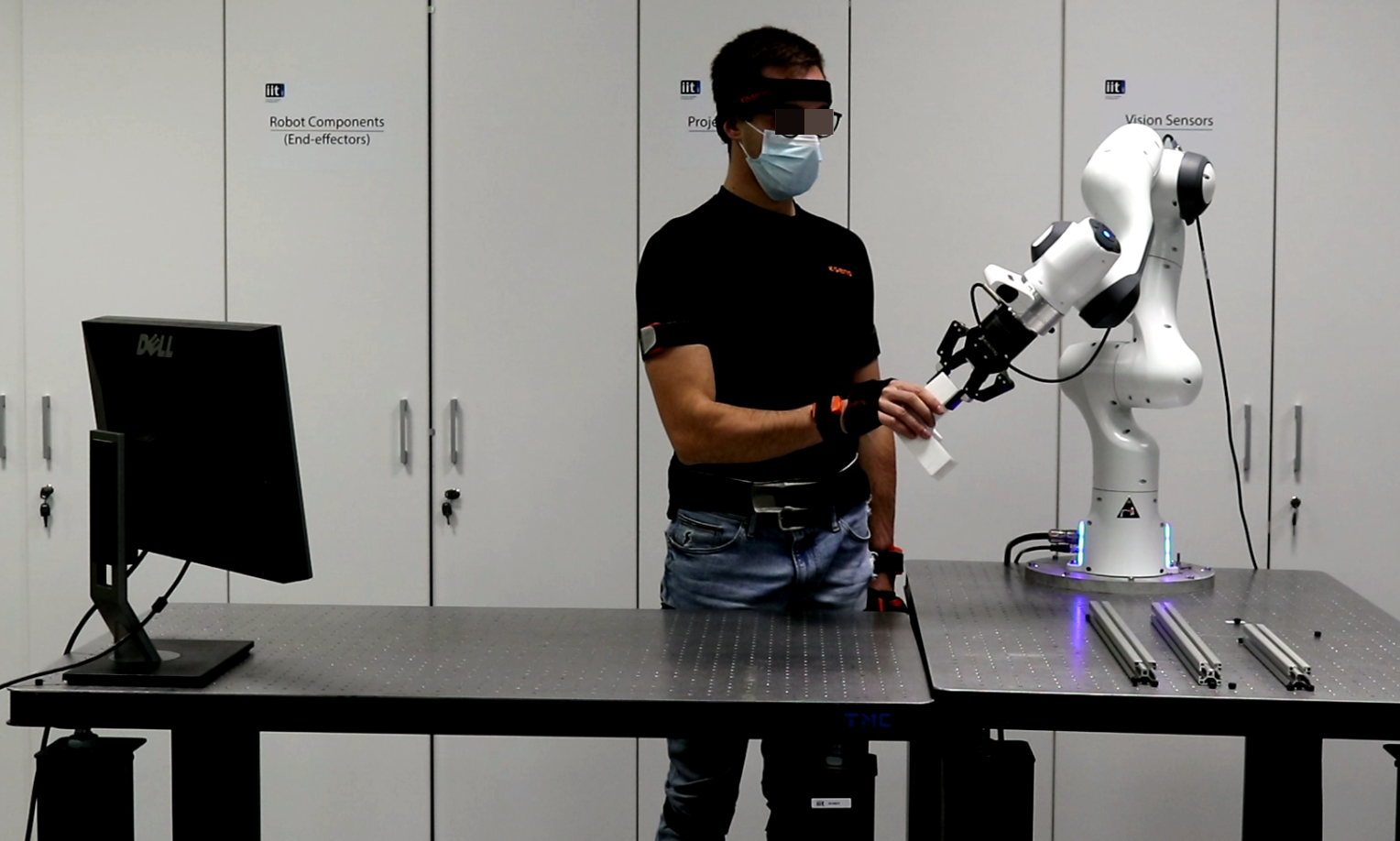}
\caption{(top) Pick and place object $J$ action assigned to the human worker. (bottom) Pick and place object $J$ action assigned to the robot.}
\label{fig:hrc_hum_action}
\vspace{-4mm}
\end{figure}

In the \textit{HRC}, the human completed the task with the robot aid according to the instructions shown on the allocation monitor (see \autoref{fig:exp_setup_3}). 
If the action was allocated to the human, they picked and placed the indicated object (see \autoref{fig:hrc_hum_action}(top)). Otherwise, the robot grasped and handed over that object to the human counterpart (see \autoref{fig:hrc_hum_action}(bottom)).

In \textit{H only}, the human had to perform the 10 repetitions without the robot help, also choosing the assembly plan by picking one piece at a time. 
To comply with the takt time constraint, the human monitored a chronometer to start the new repetition after every $T_{takt}$. 

At the end of each execution, subjects answered the NASA Task Load Index (NASA-TLX) questionnaire~\cite{hart1988development}, one per modality, which includes a section to estimate the relevance of each aspect in the completion of the task.
To evaluate if some significant difference occurs between the two task modes, we also performed the Wilcoxon signed-rank statistic test, a non-parametric statistical hypothesis test \cite{woolson2007wilcoxon}. 
Based on the number of subjects and the chosen level of significance $\alpha =0.05$, the resultant critical value of such a test is $W_{crit}=13$. Finally, we conducted a post-hoc power analysis using G* power \cite{faul2009statistical}: the rule-of-thumb is that the statistical power should be higher or equal than 80\% to ensure that the test correctly rejects the null hypothesis, i.e., there is no difference between the two modalities.

Moreover, volunteers rate a custom 7-point Likert-scale questionnaire (from "Not at all" $-3$ to "Extremely" $3$) to assess the three main aspects of our \textit{HRC} framework, i.e., the efficiency of the role allocation algorithm, the perceived ergonomics conditions, and the usability of the presented method in industrial environments. The statements to be graded follow:
\begin{enumerate}
    \item I was glad to take a break when the task was allocated to the robot.
    \item The robot was hindering me in the task accomplishment.
    \item It was easy to understand when I had to take action.
    \item When I was tired, the robot was not able to intervene.
    \item I feel safe when collaborating with the robot.
    \item The robot can correctly choose which action is too demanding for me.
    \item I would suggest using the robot for this kind of tasks.
    \item The robot help makes me feel less fatigued at the end of the task.
    \item I had to wait for the system to inform me about the next task.
\end{enumerate}

\paragraph{Results}
The results of role allocation are the following: given a total of $480$ actions to be allocated (12 subjects $\times$ 10 repetitions per subject $\times$ 4 actions per repetition), $179$ were allocated to the robot ($38\%$), while $301$ to the worker ($62\%$).
In particular, by considering each action individually, $a_1$ was allocated to the robot $51\%$ of the times, $a_2$ $78\%$, $a_3$ and $a_4$ $10\%$.

\begin{figure}[t]
\centering
	\includegraphics[width=8cm,height=5cm]{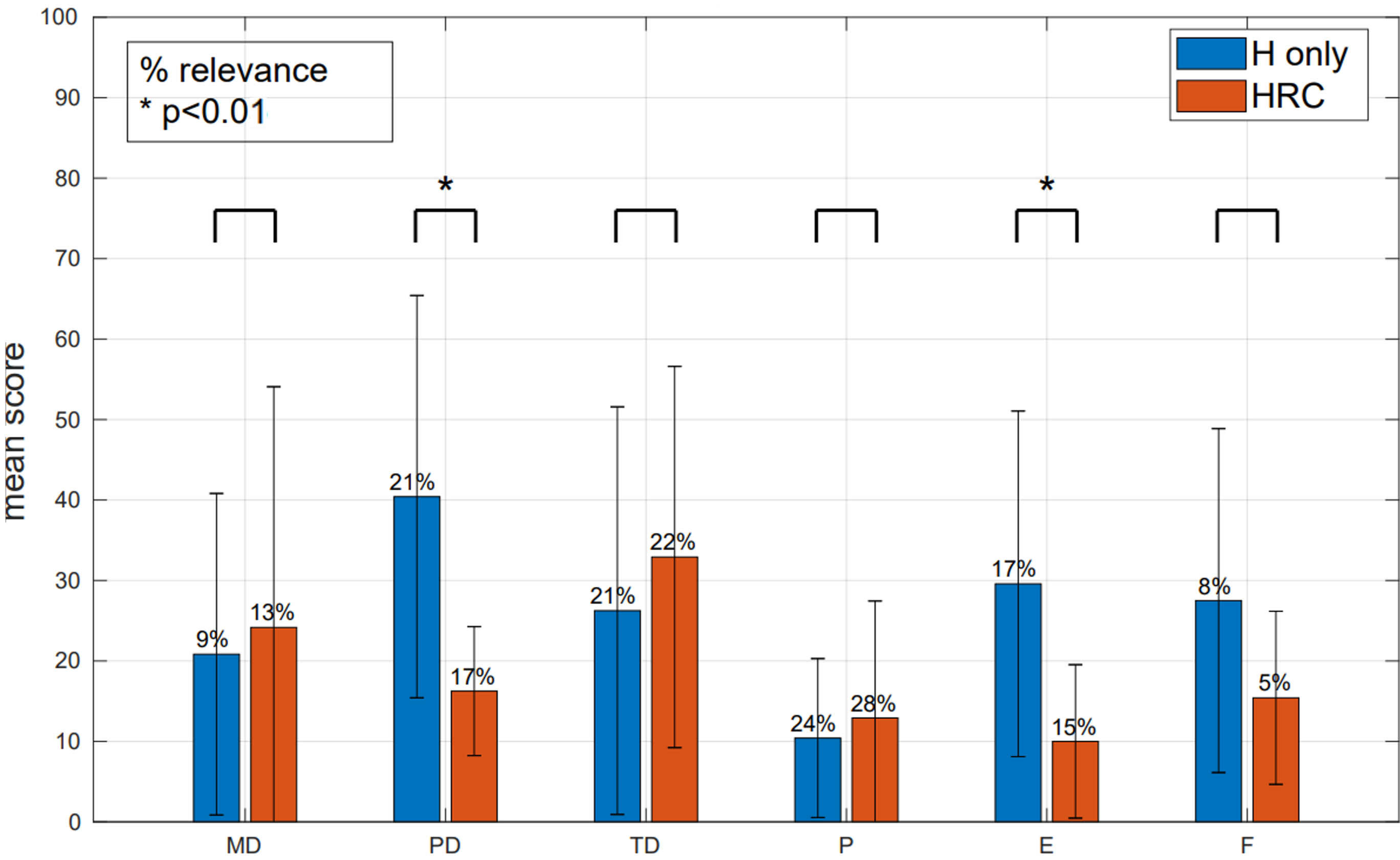}
	\caption{Results by collecting the subjects' answers to the NASA (Task Load indeX) TLX questionnaire. The column height corresponds to the mean value of the subjects' scores for each task aspect, while bars encode the standard deviation. Rated aspects are: Mental Demand (MD), Physical Demand (PD), Temporal Demand (TD), Performance (P), Effort (E), and Frustration (F). The two matched columns are for the experiment in \textit{H only} (blue one) and \textit{HRC} (red one) mode. The percentage above each column represents how much subjects consider that specific aspect relevant in completing the task.}
	\label{fig:nasa}
	\vspace{-3mm}
\end{figure}

The results of the subjects' responses to the NASA TLX questionnaire are shown in \autoref{fig:nasa}, \textit{H only} in blue, and \textit{HRC} in orange. 
The percentage above each column represents how much subjects consider that specific aspect relevant to task accomplishment. Overall, in the two experiments, these scores are comparable.

Physical Demand (PD) is higher for the \textit{H only} mode. Moreover, the difference between the two conditions was found to be statistically significant according to the Wilcoxon signed-rank test. The p-value returned by the comparison for PD in the two experimental conditions is $p<0.01$, while the test statistic is $W_{stat}=1.5$. Since $W_{stat}<W_{crit}$, we have statistically significant evidence. The statistical power is 94\%.
Concerning the Performance (P), since the task was simple, all the subjects were satisfied with their work in both conditions (the lower the bar, the higher the performance). Anyway, to reach the same level of performance, subjects retained to have worked harder for the task accomplishment in the \textit{H only} mode, as reported by the statistically significant results in the Effort (E) columns ($p<0.01$, $W_{stat}=7.5$, statistical power 78\%).
The aspects slightly penalised during the collaborative mode are Mental Demand (MD) and Temporal Demand (TD). However, the disparity is not statistically significant. In particular, a similar score for TD was expected, since we fixed the takt time $T_{takt}$ for the two experimental conditions.
Finally, the level of Frustration (F) was higher for the \textit{H only} mode. 

\begin{figure}[t]
\centering
	\includegraphics[width=8cm,height=5cm]{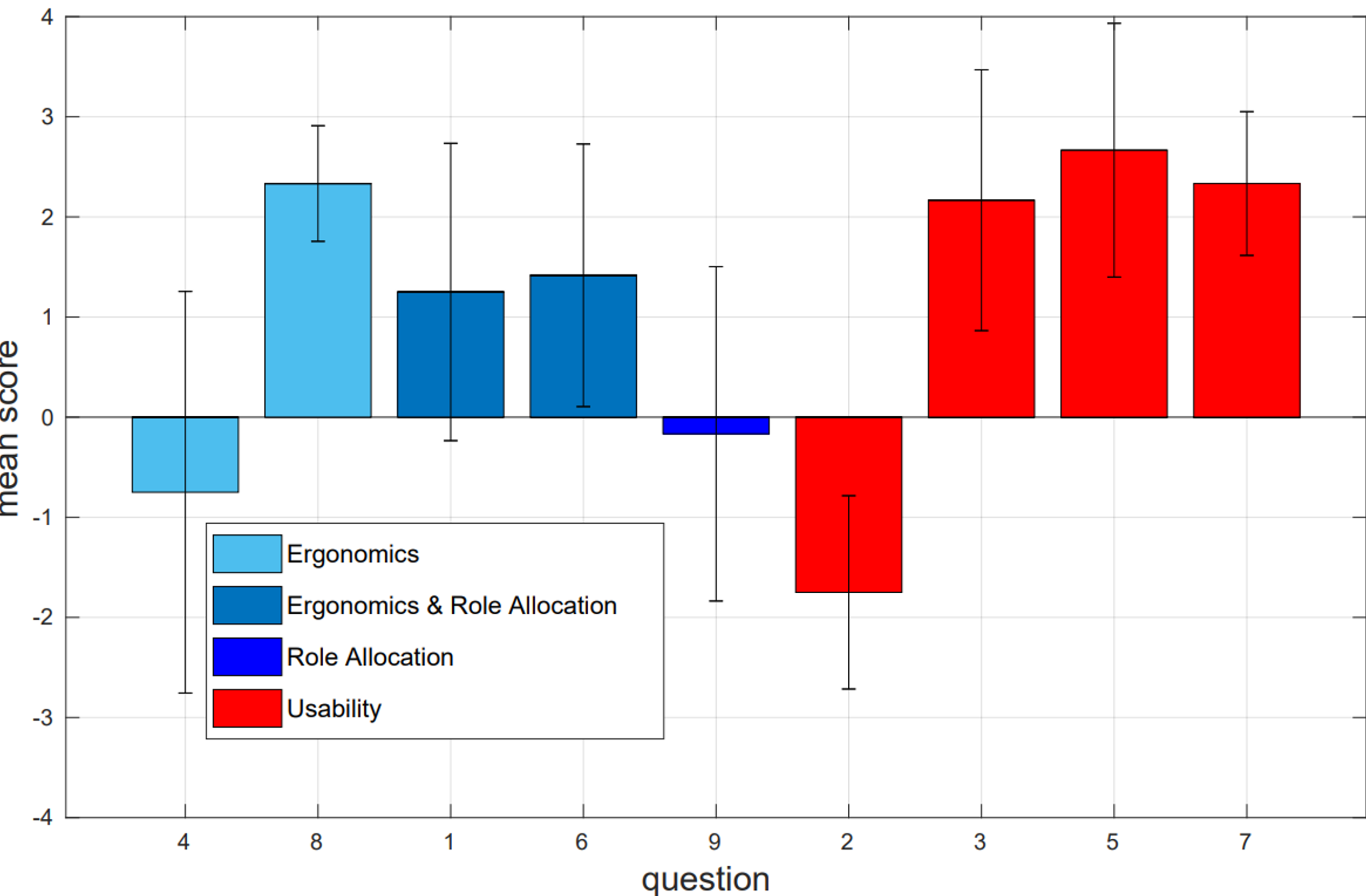}
	\caption{Results by collecting subjects' answers to the custom questionnaire. On the $x$-axis the question number, and on the $y$-axis the average score. A positive score means a positive answer and vice versa. Questions were designed to evaluate the granted Ergonomics, Role Allocation efficiency, and Usability of such a framework.}
	\label{fig:custom}
	\vspace{-4mm}
\end{figure}

The results of the custom questionnaire, in~\autoref{fig:custom}, reveal that the framework was perceived as beneficial in terms of ergonomics (questions n.1-4-6-8) and usability (questions n.2-3-5-7). Subjects did not report high waiting times due to the algorithm computational time (question n.9) or complications in task execution due to the presence of the robot (questions n.2-4-5-6-7-8). 

\paragraph{Discussion}
\label{par:7}

The statistical difference found for both Physical Demand and Effort aspects, according to which the \textit{H only} condition was rated worse, together with the recurrent allocation of the most demanding action $a_2$ ($78 \%$ of the time), suggests that our framework has the potential to improve human ergonomics. Moreover, the power analysis confirms that evaluating the performance of 12 subjects is sufficient for highlighting the differences between the two experimental modalities in terms of Physical Demand ($94\% > 80\%$) and almost enough in terms of Effort ($78\%\approx 80\%$).
Also, from the examination of the custom questionnaire results, it emerges that subjects were glad to take a break when the task was allocated to the robot. This suggests that the fatigue level at the end of the task might be lower than in completely manual operations.
A single subject felt more productive in the \textit{H only} mode due to the simplicity of the task and the limited number of repetitions. We believe that in a real industrial assembly, composed of a larger number of pieces and prolonged executions, the benefits of the proposed framework would be more significant. 

The Mental Demand factor was slightly penalised during the \textit{HRC} mode. According to subjects' experiences, the monitor-based user interface has limitations in providing intuitively the results of the allocated tasks. 
However, it could be possible to adopt more intuitive interfaces, e.g., based on Augmented Reality~\cite{defranco2019intuitive}. In any case, how to communicate the allocation results to humans falls beyond the primary objective of the work.

The level of Frustration, on the contrary, decreases with the robot assistance. In our opinion, this is also related to the task timing requirements. During the \textit{HRC} mode, the assembly rhythm was marked by the robot executions, while, without any co-worker, when the human finished the repetition, he had to wait until $T_{takt}$ passed, to start a new repetition. This wait could irritate or annoy subjects. 
On the other hand, as expected, working at a steady rhythm generates less frustration. 
Finally, the computational results, together with subject perception about the reactivity of the system, show that the algorithm does not introduce waiting times in the task execution so that productivity levels remain high.

\section{Conclusions}
\label{sec:concl}
In this manuscript, we proposed a dynamic human-robot role allocation method for industrial assembly tasks that integrates an online evaluation of the ergonomic risk. By intervening in the role distribution within the human-robot team, we aim at exploiting the robot to prevent workers' potential harm in the repetitive execution of hard tasks.

An adapted AOG structure is used for modelling a collaborative assembly task by introducing, for each action, as many hyper-arcs as the number of workers involved.
All the hyper-arcs are also characterised by a cost that encodes the suitability of agents to the modelled actions. Our framework provides for the dynamic updating of such costs according to the human state changes during task performance. An optimal allocation solution is returned by an AO* search. The algorithm presents exponential complexity in the worst case, but some rules to drastically reduce the task model connections were discussed, obtaining quasi-linear complexity. 
With such simplifications, the computational time necessary to AO* for returning a solution should allow the workers to perform the task without any interruption of the workflow, in the case of a general industrial assembly. 
A drawback of the AOG-based task modelling consists of the impossibility of introducing parallelism, i.e., the allocation algorithm cannot return as a solution two (or many) actions to be executed at the same time by the two (or many) agents involved. 

While the framework is general and envisions the possibility to employ different ergonomic indicators, in this work Kinematic Wear was designed and exploited. Unlike most of the traditional ergonomic indexes, it can model the history of the human worker's joint wear and predict future ones, using a suitable calibration procedure. Although this kinematic index has been proven a promising approach for evaluating the ergonomic risk over time, it needs to be validated with the support of physiological and clinical data, which will be the focus of successive works.  

Nevertheless, the promising results of the multi-subject experiment showed the technological acceptability and the potential of the framework in improving ergonomics during repetitive industrial assemblies.
Future improvements will introduce parallelism in the task model and the possibility for the human worker to negotiate with the robot on the allocation.

\section*{Declaration of competing interest}
The authors declare that they have no known competing financial
interests or personal relationships that could have appeared to influence
the work reported in this paper.

\section*{Acknowledgment}
This work was supported in part by the European Research Council's (ERC) starting grant Ergo-Lean (GA 850932) and in part by the European Union’s Horizon 2020 research and innovation program CONCERT (GA 101016007).

\bibliographystyle{ieeetr}
\balance
\bibliography{biblio}

\end{document}